\documentclass[11pt,letterpaper]{article}
\usepackage{caption}

\usepackage{hhline}
\usepackage{makecell}
\usepackage{subcaption}
\usepackage{fullpage}
\usepackage{times}
\usepackage{array}
\usepackage{ragged2e}
\usepackage{multirow}
\usepackage{fancyhdr, amssymb}
\usepackage[ruled,vlined]{algorithm2e}
\usepackage[dvipsnames]{xcolor}
\usepackage[margin = 1 in]{geometry}
\usepackage{graphicx}
\usepackage{outlines}
\usepackage{longtable}
\usepackage{amsmath}

\usepackage{graphicx}
\usepackage{url}
\usepackage[english]{babel} 
\usepackage{blindtext}
\usepackage{xltabular}
\usepackage{arydshln} 
\usepackage{wrapfig}
\usepackage{enumitem}
\usepackage{floatrow}
\usepackage{xfrac}
\usepackage[font={footnotesize}, labelfont={bf}, labelsep=space, justification=justified, skip=0pt]{caption}
\usepackage[hang,flushmargin]{footmisc} 
\usepackage[colorlinks=true, allcolors=blue]{hyperref}
\usepackage{csquotes}
\usepackage{amssymb}
\usepackage{bbm}
\usepackage[
backend=biber,
style=numeric-comp,
sorting=none,
maxnames = 10
]{biblatex}
\makeatletter
\AtBeginDocument{%
  \g@addto@macro\normalsize{%
    \setlength\abovedisplayskip{8pt}
    \setlength\belowdisplayskip{8pt}
    \setlength\abovedisplayshortskip{8pt}
    \setlength\belowdisplayshortskip{8pt}
    \let\orig@setfontsize\@setfontsize
  }%
}
\makeatother

\DeclareMathSizes{10}{10}{7}{6}

\newcommand{\cmt}[1]{} 

\newcommand{\fbest}{$y^*$}

\newcommand{\tb}{$\boldsymbol{t}$}

\newcommand{\borehole}{\texttt{Borehole}}
\newcommand{\wing}{\texttt{Wing}}
\newcommand{\nta}{\texttt{NTA}}
\newcommand{\hoip}{\texttt{HOIP}}
\newcommand{\mfbo}{\texttt{MFBO}}
\newcommand{\mfbonew}{\texttt{MFBO\textsubscript{UQ}}}

\title{\fontsize{14}{14}\selectfont On the Effects of Heterogeneous Errors on Multi-fidelity Bayesian Optimization}

\date{\vspace{-5ex}}
\usepackage{authblk}
\author[1]{Zahra Zanjani Foumani}
\author[1]{Amin Yousefpour}
\author[1]{Mehdi Shishehbor}
\author[1]{Ramin Bostanabad \thanks{\noindent Corresponding Author: Raminb@uci.edu}}
\affil[1]{Department of Mechanical and Aerospace Engineering, University of California, Irvine}

\begin{document}
    \include{pythonlisting}
    \pagenumbering{arabic}
    \sloppy
    \maketitle
    \noindent \textbf{Abstract}\\
Bayesian optimization (BO) is a sequential optimization strategy that is increasingly employed in a wide range of areas including materials design. In real world applications, acquiring high-fidelity (HF) data through physical experiments or HF simulations is the major cost component of BO. To alleviate this bottleneck, multi-fidelity (MF) methods are used to forgo the sole reliance on the expensive HF data and reduce the sampling costs by querying inexpensive low-fidelity (LF) sources whose data are correlated with HF samples. However, existing multi-fidelity BO (MFBO) methods operate under the following two assumptions that rarely hold in practical applications: $(1)$ LF sources provide data that are well correlated with the HF data on a global scale, and $(2)$ a single random process can model the noise in the fused data. These assumptions dramatically reduce the performance of MFBO when LF sources are only locally correlated with the HF source or when the noise variance varies across the data sources. In this paper, we dispense with these incorrect assumptions by proposing an MF emulation method that $(1)$ learns a noise model for each data source, and $(2)$ enables MFBO to leverage highly biased LF sources which are only locally correlated with the HF source. We illustrate the performance of our method through analytical examples and engineering problems on materials design.

\noindent \textbf{Keywords:} Bayesian optimization; multi-fidelity modeling; emulation; Gaussian process; interval score.
    \section{Introduction} \label{Sec: intro}

Bayesian optimization (BO) is a sequential and sample-efficient global optimization technique that is increasingly used in the optimization of expensive-to-evaluate (and typically black-box) functions \cite{shahriari2015taking}. BO has two main ingredients: an emulator which is typically a Gaussian process (GP) and an acquisition function (AF) \cite{frazier2018tutorial}. The first step in BO is to train an emulator on some initial data. Then, an auxiliary optimization is solved to determine the new sample that should be added to the training data. The objective function of this auxiliary optimization is the AF whose evaluation relies on the emulator. Given the new sample, the training data is updated and the entire emulation-sampling process is repeated until the convergence conditions are met \cite{li2020multi}. 

Although BO is a highly efficient technique, the total cost of optimization can be substantial if it solely relies on the accurate but expensive high-fidelity (HF) data source. To mitigate this issue, multi-fidelity (MF) techniques are widely adopted \cite{song2019general, takeno2020multi, zhang2019efficient} where one uses multiple data sources of varying levels of accuracy and cost in BO. The fundamental principle behind MF techniques is to exploit the correlation between low-fidelity (LF) and HF data to decrease the overall sampling costs \cite{shu2021multi, tran2020smf}.

Over the past two decades many multi-fidelity BO (MFBO) strategies have been proposed which primarily differ in terms of their emulator and AF. Most existing strategies rely on the Co-Kriging method \cite{xiao2018extended}, Kennedy and O’Hagan’s bi-fidelity approach \cite{kennedy2001bayesian}, and the BoTorch package \cite{gardner2018gpytorch}. These MFBO methods have some major drawbacks such as inability to simultaneously leverage multiple LF sources, sensitivity to the sampling costs (where highly inexpensive LF sources are heavily sampled which, in turn, causes numerical and convergence issues), and presuming simple bias forms (e.g., an additive function \cite{kennedy2001bayesian}) for the LF sources. 

Some of these limitations are recently addressed in \cite{foumani2023multi} where the authors propose to (1) use latent map Gaussian processes (LMGPs) for emulation, and (2) quantify the information value of LF and HF samples differently. Their AF is cost-aware in that it considers the sampling cost in quantifying the value of HF and LF data points. Henceforth, we refer to this method as \mfbo.

\begin{figure*}[!b] 
    \centering
    \begin{subfigure}{.5\textwidth}
        \centering
        \includegraphics[width=1\linewidth]{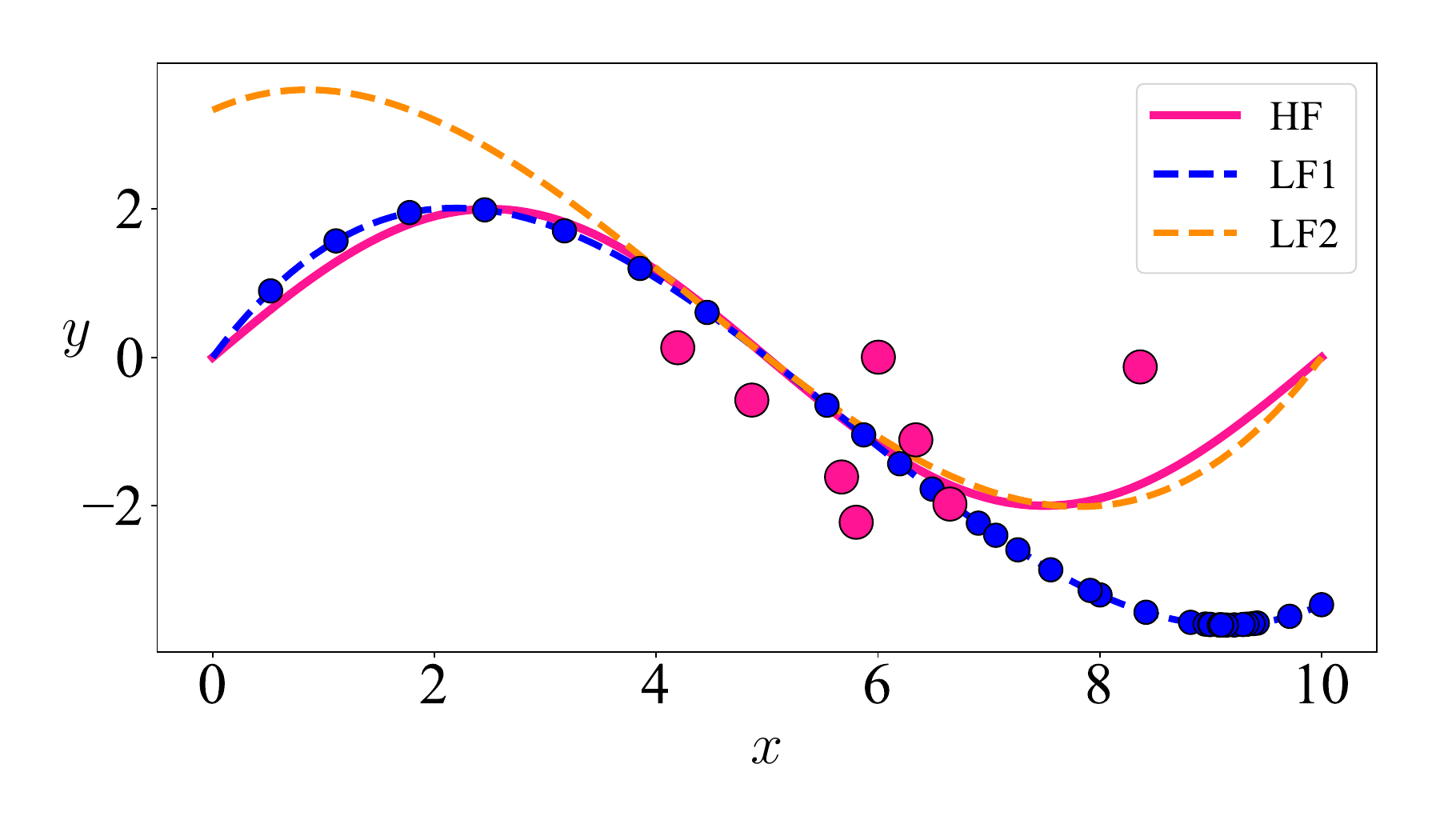}
        \vspace{-9mm}
        \caption{\textbf{\mfbo}}
        \label{fig: 1d_alpha}
    \end{subfigure}%
   \begin{subfigure}{.5\textwidth}
        \centering
        \includegraphics[width=1\linewidth]{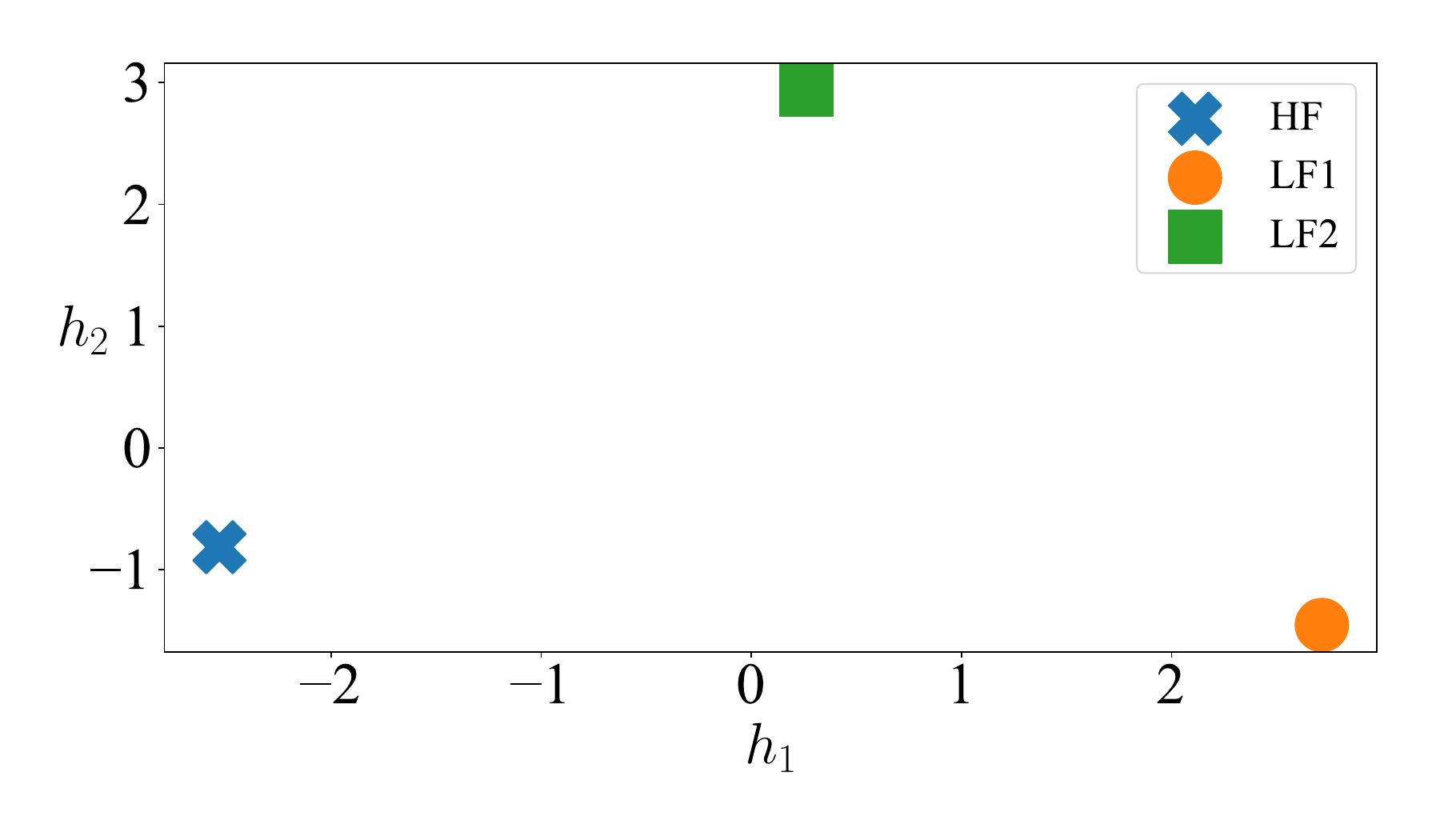}
        \vspace{-9mm}
        \caption{\textbf{Fidelity Manifold}}
        \label{fig: 1d_LM}
    \end{subfigure}%
    \newline
     \begin{subfigure}{.494\textwidth}
        \centering
        \includegraphics[width=1\linewidth]{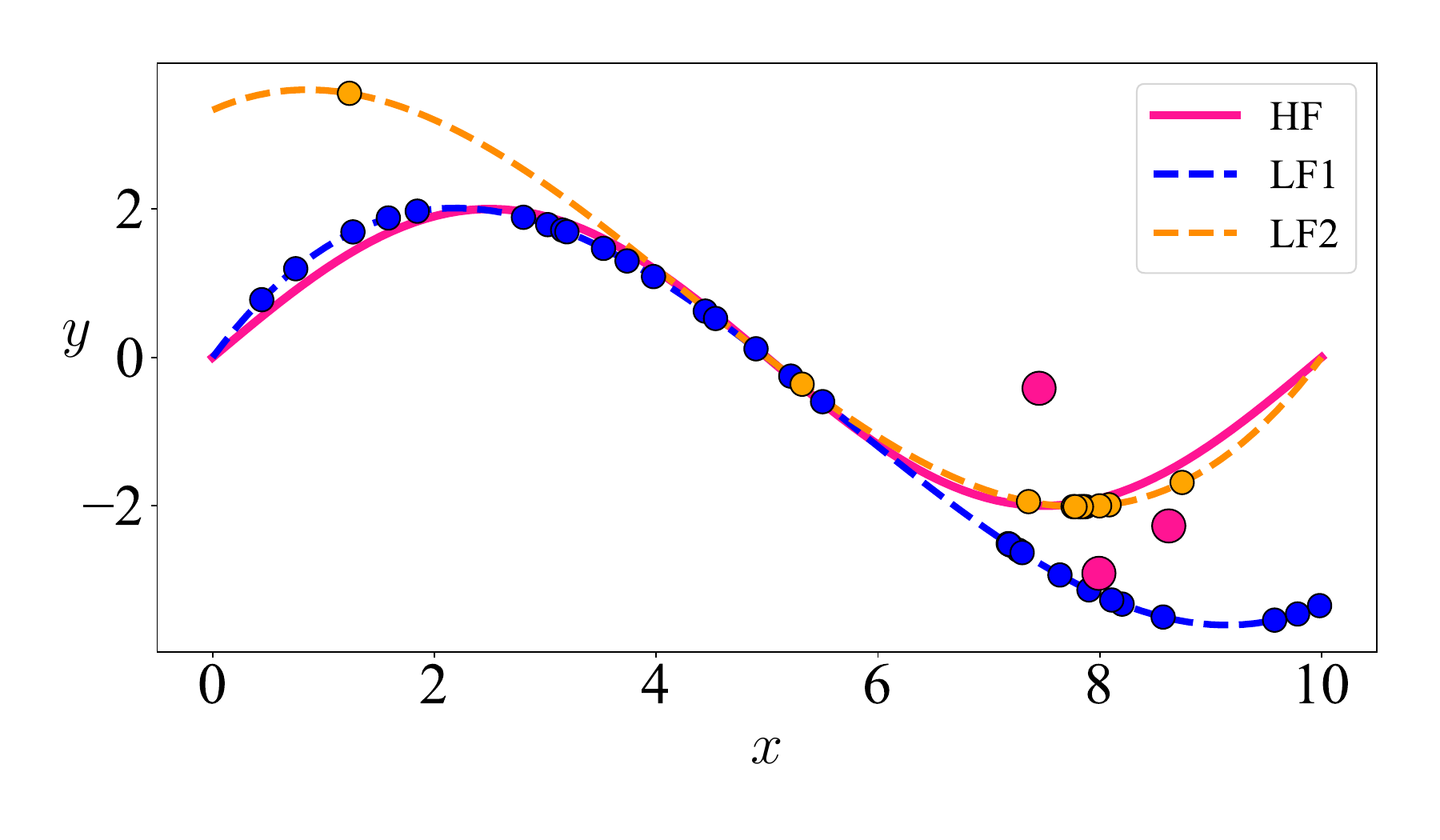}
        \vspace{-9mm}
        \caption{\textbf{\mfbonew}}
        \label{fig: 1d_beta}
    \end{subfigure}
    \begin{subfigure}{.494\textwidth}
        \centering
        \includegraphics[width=1\linewidth]{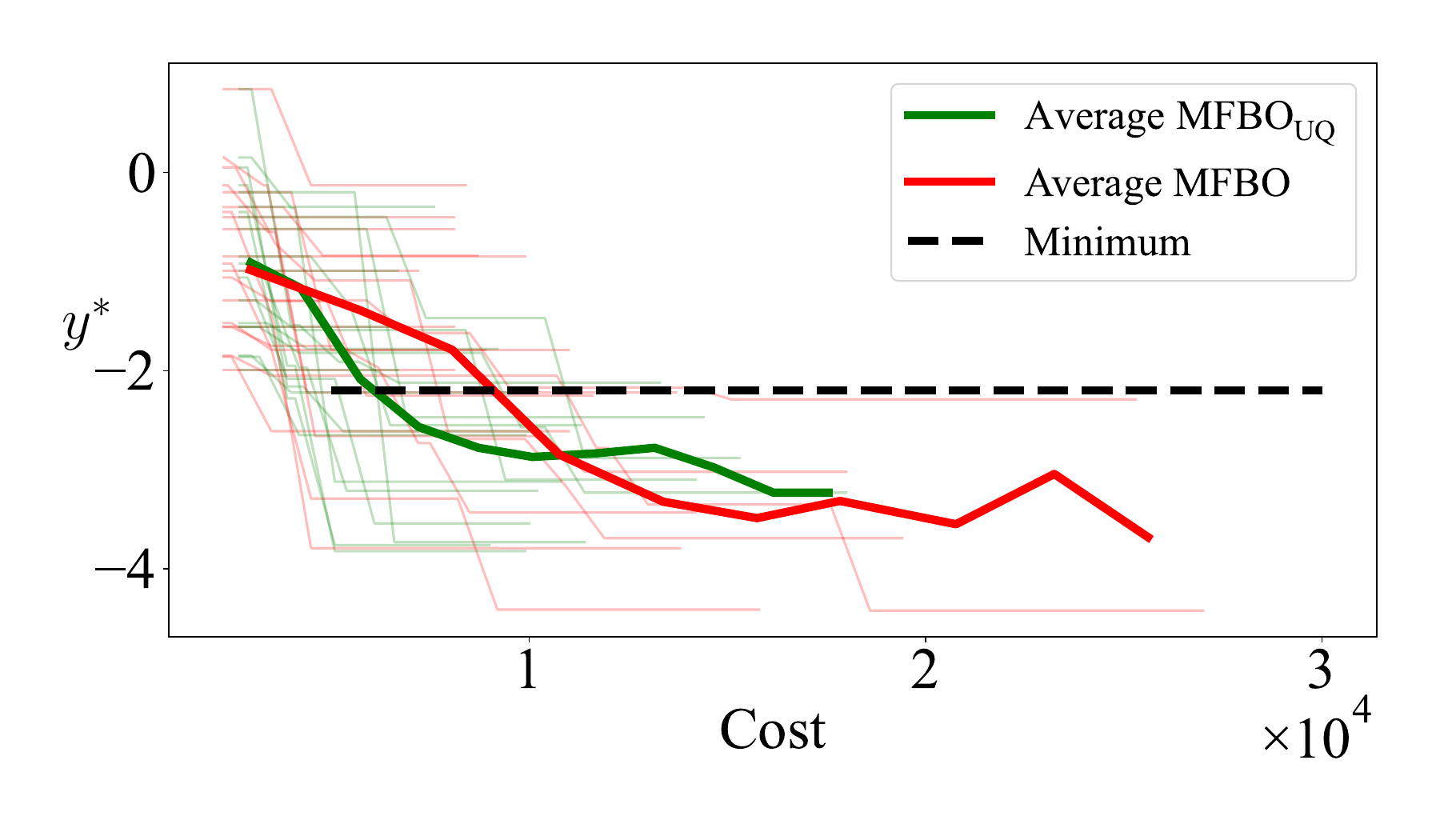}
        \vspace{-9mm}
        \caption{\textbf{Convergence History}}
        \label{fig: 1d_convergence}
    \end{subfigure}
    \caption{\textbf{Effect of heterogeneous noise and model fidelity on MFBO:} { HF data are noisy ($\sigma_{noise}=1$) and expensive while the LF data are deterministic and cheap. In this example, LF1 is more correlated with the HF source for $x<5$ while LF2 correlates better with HF data for $x>5$. The sampling cost of the HF and two LF sources are $10/1/1$, respectively. 
    \textbf{(a)} demonstrates the sampling history of \mfbo~which excludes LF2 from the sampling process based on the fidelity manifold of LMGP shown in \textbf{(b)} which indicates that LF2 is globally more correlated with the HF data compared to LF1. The emulator in \textbf{(a)} learns a single noise process for LF2 and HF data and the manifold in \textbf{(b)} is learnt via the initial data.
    \textbf{(c)} \mfbonew~is the approach we propose in this paper effectively explores the space (as it samples the HF source more in $x>5$) and leverages LF1 in  $x<5$. As for LF2, \mfbonew~mostly samples from $x>5$ since this region includes the optimum of LF2 and is more correlated with the HF. 
    \textbf{(d)} \mfbonew~ outperforms \mfbo~ in finding the optimum of HF source (\fbest) for various initial conditions (the large noise variance of HF data cause both approaches to have some errors upon convergence). 
    Initial data are not shown \textbf{(a)} and \textbf{(c)}.}}
    \label{fig: 1d}
\end{figure*}

While \mfbo~ performs much better than competing MF approaches, it has two main limitations which are demonstrated with a simple $1D$ example in Figure \ref{fig: 1d} where each of the two LF sources is more correlated with the HF function in half of the domain, see Figure \ref{fig: 1d}\textbf{a}. 
Firstly, \mfbo~ excludes highly biased LF sources from BO with the rationale that they can steer the search process in the wrong direction. This exclusion is done before BO starts since the decision is made based on the fidelity manifold of LMGP that is trained on the initial data, see Figure \ref{fig: 1d}\textbf{b}. In this manifold each data source is encoded with a point and hence distances between these points correspond to \textit{global} correlations between the corresponding data sources. That is, Figure \ref{fig: 1d}\textbf{b} suggests that the HF source is barely correlated with the LF sources even though they are close to the HF function in half of the domain (since eliminating both sources convert the BO into a single-fidelity process in this example, we assume that \mfbo~ only excludes LF2 and hence LF1 and the HF sources are sampled via \mfbo~ in Figure \ref{fig: 1d}\textbf{a}). This is obviously a sub-optimal decision as it precludes the possibility of leveraging an LF source that is valuable in a small portion of the search space which may include the global optimum of the HF source. That is, in the example of Figure \ref{fig: 1d}\textbf{a} \mfbo~ should ideally leverage LF2 but \textit{mostly} sample from it in the $x > 5$ region. 


The second limitation of \mfbo~is that it assumes all sources are corrupted with the same noise process (with unknown noise variance). However, MF datasets typically have different levels of noise especially if some sources represent deterministic computer simulations while others are physical experiments \cite{escamilla2003hybrid, kreibich2013quality}. In such applications, \mfbo~overestimates the uncertainties associated with the noise-free data sources which, in turn, reduces the performance of MFBO. 

To address these two limitations, we introduce \mfbonew~for multi-fidelity cost-aware Bayesian optimization. \mfbonew~ has the same AFs as \mfbo~and can leverage an arbitrary number of LF sources in optimizing an HF source. Unlike \mfbo, \mfbonew~never discards an LF source (regardless of its bias with respect to the HF source) and estimates a noise process for each data source. We argue that \mfbonew~quantifies the uncertainties more accurately than \mfbo~and thus achieves a higher performance in MFBO.  
Figure \ref{fig: 1d}\textbf{c} schematically demonstrates the advantages of \mfbonew~over \mfbo~ in a $1D$ example where there are one HF and two LF sources. As it can be observed the LF sources are mostly sampled where they $(1)$ are well correlated with the HF source, or $(2)$ provide attractive function values (e.g., very small values in minimization). The advantages of \mfbonew~ over \mfbo~ in finding the optimum of HF (\fbest) hold over various initializations, see Figure \ref{fig: 1d}\textbf{d}.

The rest of the paper is organized as follows. We provide the methodological details in Section \ref{Sec: Method} and then evaluate the performance of \mfbonew~via multiple ablation studies in Section \ref{Sec: results}. In Section \ref{Sec: results}, we also visualize how strategic sampling in \mfbonew, driven by accurate uncertainty quantification and effective handling of biased data sources, results in its superior performance compared to \mfbo. This is demonstrated through two real-world high-dimensional material design examples with noisy and highly biased data sources. We conclude the paper in Section \ref{Sec: conclusion} by summarizing our contributions and providing future research directions.

    \section{Methods} \label{Sec: Method}
In this section, we first provide  some background on LMGP and MF modeling with LMGP in Section \ref{Sec: LMGP} and Section \ref{Sec: MF-LMGP}, respectively. We then propose our efficient mechanism for inversely learning a noise process for each data source in Section \ref{Sec: multiple-noise}. Next, we introduce the cost-aware AF of \mfbonew~in Section \ref{Sec: CA-AF}. Finally, in Section \ref{Sec: Interval_score} we elaborate on our idea that improves the uncertainty quantification (UQ) capabilities of LMGPs and, in turn, benefits MFBO. 

\subsection{Latent Map Gaussian Process (LMGP)} \label{Sec: LMGP}
Gaussian processes (GPs) are emulators which assume the responses or outputs in the training data come from a multivariate normal distribution with parametric mean and covariance functions that depend on the inputs. Based on this assumption, the following equation can be written:
\begin{equation} 
    \begin{split}
        y(\boldsymbol{x})=\beta+\xi(\boldsymbol{x})
    \end{split}
    \label{eq: GP-prior}
\end{equation}
\noindent where $\boldsymbol{x}=[x_1,x_2,\dots,x_{dx}]^T$ is the input vector, $y(\boldsymbol{x})$ is the output, $\beta$ is an unknown coefficient, and $\xi(\boldsymbol{x})$ is a zero-mean GP with the covariance function:
\begin{equation} 
    \begin{split}
        cov(\ \xi(\boldsymbol{x}),\ \xi(\boldsymbol{x}^\prime))=c(\boldsymbol{x},\boldsymbol{x}^\prime)=\ \sigma^2r(\boldsymbol{x},\boldsymbol{x}^\prime)
    \end{split}
    \label{eq: GP-Cov}
\end{equation}
\noindent where $\sigma^2$ is the variance of the process and $r(\cdot,\cdot)$ is the parametric correlation function which measures the distance between any two input vectors. In this paper we use the Gaussian correlation function defined as:
\begin{equation} 
    \begin{split}
        r(\boldsymbol{x},\boldsymbol{x}^\prime)=\exp{\{-\sum_{i=1}^{dx}{10}^{\omega_i}\ (x_i-\ x_i^\prime)^2\}}
    \end{split}
    \label{eq: GP-correlation}
\end{equation}
\noindent where $\boldsymbol{\omega}=[\omega_1, \omega_2 ,\dots,\omega_{dx}]^T$ are the scale parameters. The versatility of a GP emulator highly depends on its correlation function in that, e.g., traditional GPs do not accommodate categorical variables directly since kernels like the one in Equation \ref{eq: GP-correlation} cannot compare qualitative features. 

To directly use GPs in MF modeling, we follow \cite{eweis2022data} who convert MF modeling to a manifold learning problem via LMGPs which are extensions of GPs that can handle categorical data \cite{oune2021latent} while providing a visualizable manifold that can be used to interpret the correlation among data sources. 

Denoting the categorical inputs by $\boldsymbol{t}=[t_1,t_2,\dots,t_{dt}]^T$ where variable $t_i$ has $l_i$ distinct levels, LMGP maps each combination of the categorical levels to a point in a learned quantitative manifold. To this end, LMGP assigns a unique vector to each combination of the categorical variables and then uses a parametric function to map these unique vectors into a compact manifold with dimensionality $dz$. Assuming a linear transformation is used in LMGP, the mapping operation reads as:
\begin{equation} 
    \begin{split}
        \boldsymbol{z}(\boldsymbol{t})=\boldsymbol{\zeta}(\boldsymbol{t})\boldsymbol{A}
    \end{split}
    \label{eq: GP-mapping}
\end{equation}
\noindent where $\boldsymbol{t}$ denotes a specific combination of the categorical variables, $\boldsymbol{z}(\boldsymbol{t})$ is the $1\times dz$ posterior latent representation of $\boldsymbol{t}$, $\boldsymbol{\zeta}(\boldsymbol{t})$ is a unique prior vector representation of $\boldsymbol{t}$, and $\boldsymbol{A}$ is a rectangular matrix that maps $\boldsymbol{\zeta}(\boldsymbol{t})$ to $\boldsymbol{z}(\boldsymbol{t})$. In this paper, grouped one-hot encoding is used to generate the prior vectors and hence the dimensionality of $\boldsymbol{\zeta}(\boldsymbol{t})$ and $\boldsymbol{A}$ are $1\times \sum_{i=1}^{dt}l_i$ and $\sum_{i=1}^{dt}l_i \times dz$, respectively. These mapped points can now be directly embedded in the correlation function as: 
\begin{equation} 
    \begin{split}
        r(\boldsymbol{u},\boldsymbol{u}^\prime)=\exp{\{-\sum_{i=1}^{dx}{10}^{\omega_i}\ (x_i-\ x_i^\prime)^2\}}
        \sum\{-\sum_{i=1}^{dz}(z_i(\boldsymbol{t})-\ z_i(\boldsymbol{t}^\prime))^2\}
    \end{split}
    \label{eq: LMGP-correlation}
\end{equation}
\noindent where $\boldsymbol{u}=[\boldsymbol{x};\ \boldsymbol{t}]$ and $\boldsymbol{z}(\boldsymbol{t})=[z_1(\boldsymbol{t}),\ z_2(\boldsymbol{t}),\ \ldots,\ z_{dz}(\boldsymbol{t})]$ is the location in the learned latent space corresponding to the specific combination of the categorical variables denoted by $\boldsymbol{t}$.

LMGP estimates the hyperparameters $(\beta,\ \boldsymbol{A},\ \boldsymbol{\omega},\ \sigma^2)$ via maximum a posteriori (MAP) which, assuming $dz=2$, provides point-estimates for $dx+2\ \times\sum_{i=1}^{dt}l_i$ variables. Upon parameter estimation, LMGP uses the conditional distribution formulas to predict the response distribution at the arbitrary point $\boldsymbol{u}$ with the following mean and variance:
\begin{equation} 
    \begin{split}
        \mathbb{E}[y(\boldsymbol{u})]=\mu(\boldsymbol{u})=\ \hat{\beta}+\ \boldsymbol{r}^T(\boldsymbol{u})\ \boldsymbol{R}^{-1}(\boldsymbol{y}-\boldsymbol{1}_{n\times1}\hat{\beta})
    \end{split}
    \label{eq: LMGP-mean}
\end{equation}
\begin{equation} 
    \begin{split}
        c(y(\boldsymbol{u}),y(\boldsymbol{u}))=\ \sigma^2(\boldsymbol{u})={\hat{\sigma}}^2(1-\boldsymbol{r}^T(\boldsymbol{u})\boldsymbol{R}^{-1}\boldsymbol{r}(\boldsymbol{u})+{(g(\boldsymbol{u}))}^2({\boldsymbol{1}_{1\times n}\boldsymbol{R}}^{-1}\boldsymbol{1}_{n\times1})^{-1})
    \end{split}
    \label{eq: LMGP-variance}
\end{equation}
\noindent where $n$ is number of training samples, $\mathbb{E}$ denotes expectation, $\boldsymbol{1}_{a\times b}$ is an $a\times b$ matrix of ones, $\boldsymbol{r}(\boldsymbol{u})$ is an $n\times 1$ vector with the $i^{th}$ element $r(\boldsymbol{u}^i,\boldsymbol{u})$, $\boldsymbol{R}$ is an $n\times n$ matrix with $R_{ij}=r(\boldsymbol{u}^i,\boldsymbol{u}^j)$, and~ $g(\boldsymbol{u})=1-\mathbf{1}_{1 \times n} \boldsymbol{R}^{-1} \boldsymbol{r}(\boldsymbol{u})$. 

\subsection{ Multi-fidelity Emulation via LMGP} \label{Sec: MF-LMGP}
The first step to MF emulation with LMGP is to augment the inputs with the additional \textit{categorical} variable $s$ that indicates the source of a sample, i.e., $s=\{'1','2',\ \ldots,'ds'\}$ where the $j^{th}$ element corresponds to source $j$ for $j=1,\ldots,ds$. Subsequently, the training data from all sources are concatenated and used in LMGP to build an MF emulator. Upon training, to predict the objective value of a point $\boldsymbol{x}$ from source $j$, $\boldsymbol{x}$ is concatenated with the categorical variable $s$ that corresponds to source $j$ and fed into the trained LMGP. We refer the readers to \cite{foumani2023multi} for more detail but note here that in case the input variables already contain some categorical features (see Section \ref{Sec: real_world_examples} for an example), we endow LMGP with two manifolds where one encodes the fidelity variable $s$ while the other manifold encodes the rest of the categorical variables. While this choice does not noticeably affec the accuracy of LMGP during test time, it increase interpretability. For instance, we use the learned manifold for the categorical variables in Section \ref{Sec: real_world_examples} to show the trajectory of BO in the design space.

It has been recently shown \cite{oune2021latent} that LMGPs have the following primary advantages over other MF emulators: (1) they provide a more flexible and accurate mechanism to build MF emulators since they learn the relations between the sources in a nonlinear manifold, (2) they learn all the sources quite accurately rather than just emulating the HF source, and (3) they provide a visualizable global metric for comparing the relative discrepancies/similarities among the data sources.

\subsection{Source-dependent Noise Modeling} \label{Sec: multiple-noise}
The presence of noise significantly affects the performance of BO and incorrectly modeling it can cause over-exploration or under-exploration of the search space. To mitigate the effects of noise in BO, we reformulate LMGPs to independently model a noise process for each data source. This reformulation can improve the accuracy of the model in noisy regions and, in turn, guide the search toward the global optimum when the modeled is deployed in MFBO.

To model noise in GPs, the nugget or jitter parameter, $\delta$, is used \cite{bostanabad2018leveraging} to replace $\boldsymbol{R}$ with $\boldsymbol{R}_\delta=\boldsymbol{R}+\delta \boldsymbol{I}$ where $\boldsymbol{I}$ is an $n \times n$ identity matrix. With this approach, the estimated stationary noise variance in the data is $\delta \sigma^2$ and the mean and variance formulations in Eq. \ref{eq: LMGP-mean} and Eq. \ref{eq: LMGP-variance} are modified by using $\boldsymbol{R}_\delta$  instead of $\boldsymbol{R}$. 

Although incorporating this modification in the correlation matrix can enhance the performance of the emulator and BO in single-fidelity (SF) problems, it does not yield the same benefits in MF optimization. This is likely because of the dissimilar nature of the data sources and their corresponding noises. When dealing with multiple sources of data, each source may suffer from different levels and types of noise. Consider a bi-fidelity dataset where the HF data comes from an experimental setup and is subject to measurement noise, while the LF data is generated by a deterministic computer code which has a systematic bias due to missing physics. In this case, using only one nugget parameter in LMGP for MF emulation is obviously not an optimum choice. 

To address this issue effectively, we propose to use multiple nugget parameters in the emulator. Specifically, we define the nugget vector $\boldsymbol{\delta}=[\delta_1,\delta_2,\dots,\delta_{ds}]$ and update the correlation matrix as follows:
\begin{equation} 
    \begin{split}
        \boldsymbol{R}_\delta=\boldsymbol{R}+\boldsymbol{N}_\delta
    \end{split}
    \label{eq: sep_noise_corr}
\end{equation}
\noindent where $\boldsymbol{N}_\delta$ denotes an $n \times n$ diagonal matrix whose $(i,i)^{th}$ element is the nugget element corresponding to the data source of the $i^{th}$  sample. For instance, suppose the $i^{th}$  sample $(\boldsymbol{u}^i)$ is generated by source $ds$. Then, $(i,i)^{th}$ element of $\boldsymbol{N}_\delta$ is $\delta_{ds}$. Then, we use Eq. \ref{eq: sep_noise_corr} to build the correlation matrix of LMGP and jointly estimate all the parameters via MAP as:
\begin{equation} 
    \begin{split}
        [\widehat{\beta}, \hat{\sigma}, \widehat{\boldsymbol{\omega}}, \widehat{\boldsymbol{A}},\widehat{\boldsymbol{\delta}}]=\underset{\beta, \sigma^2, {\boldsymbol{\omega}}, \boldsymbol{A}}{\operatorname{argmin}} \frac{n}{2} \log (\sigma^2)+\frac{1}{2} \log (|\boldsymbol{R}_\delta|)+\frac{1}{2 \sigma^2}(\boldsymbol{y}-\boldsymbol{M} \beta)^T \boldsymbol{R}_\delta^{-1}(\boldsymbol{y}-\boldsymbol{M} \beta){+\log(\underset{\beta, \sigma^2, {\boldsymbol{\omega}}, \boldsymbol{A}, \boldsymbol{\delta}}{\operatorname{P(\cdot)}}) }
    \end{split}
    \label{eq: sep_noise_MAP}
\end{equation}
\noindent where $p(\cdot)$ is the prior of the hyperparameters. We define independent priors for each parameter where $\omega^i\sim N(-3,3)$, $\beta \sim N(0,1)$, $A^{ij} \sim N(0,3)$, $\sigma \sim LN(0,3)$\footnote{Log-Normal} , and $\delta^i \sim LHS(0,0.01)$\footnote{Log-Half-Horseshoe with zero lower bound and scale parameter $0.01$.}  \cite{carvalho2010horseshoe}. Our multi-noise approach increases the number of LMGP's hyperparameters to $dx+2 \times \sum_{i=1}^{d t} l_i+d s$.

\subsection{Multi-source Cost-aware Acquisition Function } \label{Sec: CA-AF}
The choice of AF is crucial in MFBO since it must consider the biases of LF data and source-dependent sampling costs in addition to balancing exploration and exploitation. To capture these goals, separate AFs are defined in \cite{foumani2023multi} for LF and HF sources with a focus on exploration and exploitation, respectively. 

Following the idea of proposing an AF with a focus on exploration for the LF sources, the AF of the $j^{th}$ LF source ($j \neq l$, $l$ denotes the HF source) is defined as the exploration part of the expected improvement (EI) in \mfbo:
\begin{equation} 
    \begin{split}
        \gamma_{LF}(\boldsymbol{u} ; j)=\sigma_j(\boldsymbol{u}) \phi(\frac{y_j^*-\mu_j(\boldsymbol{u})}{\sigma_j(\boldsymbol{u})})
    \end{split}
    \label{eq: AF_LF}
\end{equation}
\noindent where $y_j^*$ is the best function value obtained so far from source $j$ and $\phi(\cdot)$ denotes the probability density function (PDF) of the standard normal variable. $\sigma_j(\boldsymbol{u})$ and $\mu_j(\boldsymbol{u})$ are the standard deviation and mean, respectively, of point $\boldsymbol{u}$ from source $j$ which we estimate via :  
\begin{equation} 
    \begin{split}
        \mu(\boldsymbol{u})= 
\hat{\beta}+\boldsymbol{r}^T(\boldsymbol{u}) \boldsymbol{R}_\delta^{-1}(\boldsymbol{y}-\mathbf{1}_{n \times 1} \hat{\beta})
    \end{split}
    \label{eq: mean_after_sep_noise}
\end{equation}
\begin{equation} 
    \begin{split}
        c(y(\boldsymbol{u}), y(\boldsymbol{u}))= & \sigma^2(\boldsymbol{u}) 
 =\hat{\sigma}^2(1-\boldsymbol{r}^T(\boldsymbol{u}) \boldsymbol{R}_\delta^{-1} \boldsymbol{r}(\boldsymbol{u})+(g(\boldsymbol{u}))^2(\mathbf{1}_{1 \times n} \boldsymbol{R}_\delta^{-1} \mathbf{1}_{n \times 1})^{-1}) 
 +\hat{\delta}_j
    \end{split}
    \label{eq: var_after_sep_noise}
\end{equation}
\noindent where $g(\boldsymbol{u})=1-\mathbf{1}_{1 \times n} \boldsymbol{R}_\delta^{-1} \boldsymbol{r}(\boldsymbol{u})$ and $\hat{\delta}_j$  is the estimated nugget parameter for source $j$.   

\mfbo~utilizes improvement as the AF for the HF data source, since it is computationally efficient and emphasizes exploitation. Accordingly, \mfbonew~uses improvement for the HF source (source $l$) with the new mean calculated based on the Eq. \ref{eq: mean_after_sep_noise}:
\begin{equation} 
    \begin{split}
        \gamma_{H F}(\boldsymbol{u} ; l)=\mu_l(\boldsymbol{u})-y_l^*
    \end{split}
    \label{eq: HF_AF}
\end{equation}

In each iteration of BO, we first use the mentioned AFs to solve $ds$ auxiliary optimizations to find the candidate points with the highest acquisition value from each source. We then scale these values by the corresponding sampling costs to obtain the following composite AF : 
\begin{equation} 
    \begin{split}
    \gamma_{\mfbonew}(\boldsymbol{u}; j) = 
    \begin{cases}
        \sfrac{\gamma_{LF}(\boldsymbol{u}; j)}{O(j)} & j = [1, \cdots, ds] \And j \neq l \\
        \sfrac{\gamma_{HF}(\boldsymbol{u}; l)}{O(l)} & j = l
    \end{cases}
    \end{split}
    \label{eq: composite_AF}
\end{equation}
\noindent where $O(j)$ is the cost of acquiring one sample from source $j$. We determine the final candidate point (and the source that it should be sampled from) at for iteration $k+1$ via:
\begin{equation} 
    \begin{split}
    [\boldsymbol{u}^{k+1}, j^{k+1}]=\underset{\boldsymbol{u}, j}{\operatorname{argmax}}~ \gamma_{\mfbonew}(\boldsymbol{u} ; j)
    \end{split}
    \label{eq: auxiliary-opt}
\end{equation}

\subsection{Emulation for Exploration} \label{Sec: Interval_score}
The composite AF in Eq.\ref{eq: composite_AF} quantifies the information value of LF samples via Eq.\ref{eq: AF_LF} whose value scales with the prediction uncertainties, i.e., $\sigma(\boldsymbol{u})$. The source-dependent noise modeling of Section \ref{Sec: multiple-noise} improves LMGP’s ability in learning the uncertainty by introducing a few more hyperparameters. However, the added hyperparameters may result in overfitting and, in turn, deteriorate the predicted uncertainties \cite{gal2014distributed, mohammed2017over}. A related issue is the effect of large local biases of LF sources which can inflate the uncertainty quite substantially and, as a result, increase $\gamma_{LF}(\boldsymbol{u};j)$. This increase causes MFBO to repeatedly sample from the biased LF sources. Such repeated samplings reduce the efficiency of MFBO and may cause numerical issues (due to ill-conditioning of the covariance matrix) or even convergence to a suboptimal solution.

To address the above issues simultaneously, we argue that the training process of the emulator should increase the importance of UQ which directly affects the exploration part of MFBO. To this end, we leverage strictly proper scoring rules while training LMGPs.

Scoring rules \cite{lindley1982scoring} evaluate a probabilistic prediction by assigning a numerical score to it. The scoring rule of an emulator is (strictly) proper if matching the predicted distribution with the underlying sample distribution (uniquely) maximizes the expected score for any sample. The probabilistic nature of LMGP’s prediction motivates us to use the negatively oriented interval score (hereafter denoted by $IS$) to evaluate the UQ capabilities of LMGPs. We choose $IS$ since it is robust to outliers, rewards narrow prediction intervals, and is flexible in the choice of desired coverage levels \cite{bracher2021evaluating, mitchell2017proper}.

$IS$ is a special case of quantile prediction that penalizes the model for each observation that is not inside the $(1-v) \times 100 \%$ prediction interval. The lower $(\mathcal{L}^i)$ and upper $\mathcal{U}^i$ endpoints of this prediction interval for the $i^{th }$observation are their predictive quantiles at levels $v/2$ and $1-v/2$, respectively. So, we calculate the $IS$ as:
\begin{equation} 
    \begin{split}
        IS_v= \frac{1}{n} \sum_{i=1}^n(\mathcal{U}^i-\mathcal{L}^i)+\frac{2}{v}(\mathcal{L}^i-y(\boldsymbol{u}^i)) \mathbbm{1}~\{y(\boldsymbol{u}^i)<\mathcal{L}^i\} +\frac{2}{v}(y(\boldsymbol{u}^i)-\mathcal{U}^i) \mathbbm{1}\{y(\boldsymbol{u}^i)>\mathcal{U}^i\} 
    \end{split}
    \label{eq: IS}
\end{equation}
\noindent where $\mathbbm{1}\{\cdot\}$ is an indicator function which is $1$ if its condition holds and zero otherwise \cite{gneiting2007strictly, mora2023probabilistic}. We use $v=0.05$  ($95\%$ prediction interval), so $\mathcal{U}^i= \mu{(\boldsymbol{u}^i)}+1.96 \sigma{(\boldsymbol{u}^i )}$ and $\mathcal{L}^i= \mu{(\boldsymbol{u}^i )}-1.96 \sigma{(\boldsymbol{u}^i)}$.  

Having defined the $IS$, we now formulate the new objective function for training LMGPs where $IS_{0.05}$ is used as a penalty term during hyperparameter estimation to increase the focus on UQ. Since the effectiveness of this penalization mechanism depends on the value of the posterior, we introduce an adaptive coefficient whose magnitude depends on the posterior value. With this penalty term, we estimate the hyperparameters of LMGP via:
\begin{equation} 
    \begin{split}
     {[\hat{\beta}, \widehat{\sigma}, \widehat{\boldsymbol{\omega}}, \widehat{\boldsymbol{A}}, \widehat{\boldsymbol{\delta}}]=}  \underset{\beta, \sigma, \boldsymbol{\omega}, \boldsymbol{A}, \boldsymbol{\delta}}{\operatorname{argmin}} L_{MAP}  +\varepsilon|L_{MAP}| \times IS_{0.05}
    \end{split}
    \label{eq: IS_based_objective}
\end{equation}
\noindent where $|\cdot|$ denotes the absolute function and $\varepsilon$ is a user-defined scaling parameter. In this paper, we use $\varepsilon=0.08$ for all of our examples.

    \section{Results and Discussion} \label{Sec: results}
We demonstrate the performance of \mfbonew~on two analytic examples (see Table \ref{table: analytic-formulation} for details on functional forms, sampling costs, number of LF sources and their accuracy with respect to the HF source, and size of initial data) and two real-world problems. In each case, we compare the results against those of \mfbo~and single-fidelity BO (SFBO). While SFBO uses EI as its AF, \mfbonew~and \mfbo~use the AFs introduced in Section \ref{Sec: CA-AF}.

We assume that the cost of querying any of the data sources is much higher than the computational costs of BO (i.e., fitting LMGP and solving the auxiliary optimization problem). Therefore, we compare the methods based on their capability to identify the global optimum of the HF source and the overall data collection cost. By comparing these methods, we aim to demonstrate: $(1)$ the advantages of estimating noise process for each data source, $(2)$ that using IS improves the accuracy of LMGP and, in turn, enhances the convergence of BO (since our defined AFs highly rely on the quality of the prediction), and $(3)$ that deploying IS eliminates the need for excluding highly biased fidelity sources from BO. 

We use the same stop conditions across the three methods to clearly demonstrate the benefits of our two contributions. In particular, the optimization is stopped when either of the following happens: $(1)$ the overall sampling cost exceeds a pre-determined maximum budget, or $(2)$ the best HF sample does not change over $50$ iterations. The maximum budget for the analytical examples is $40000$ units, while it is $1000$ and $1800$ for the two real-world examples. These budgets are chosen based on the data collection costs. 

\subsection{Analytical Examples} \label{Sec: analytical_examples}

We consider two analytical examples, \wing and \borehole, whose input dimensionality is $10$ and $8$, respectively. To challenge the convergence and better illustrate the power of separate noise estimation, we only add noise to the HF data (the noise variance is defined based on the range of each function.). The added noise variance to the HF source of \wing~ and \borehole~ are $9$ and $16$, respectively. Both examples are single response and details regarding their formulation, initialization, and sampling cost is presented in Table \ref{table: analytic-formulation}). To assess the robustness of the results and quantify the effect of random initial data, we repeat the optimization process $20$ times for each example with each of the three methods (all initial data are generated via Sobol sequence). 

In each example, the relative root mean squared error (RRMSE) is calculated between LF sources and their corresponding HF source based on $1000$ samples to show the relative accuracy of the LF sources (presented in Table \ref{table: analytic-formulation}). Based on these ground truth numbers (which are \textit{not} used in BO), in the case of \borehole the source ID, true fidelity level (based on the RRMSEs), and sampling costs are not related (e.g., although the first LF source is the most expensive one, it has the least accuracy compared to the HF source). In the case of \wing, however, these numbers match (e.g., LF1 is the most accurate and expensive LF source and is followed by LF2 and then LF3). 

\begin{figure}[!t]
    \centering
    \begin{subfigure}{.5\textwidth}
        \centering
        \includegraphics[width=1\linewidth]{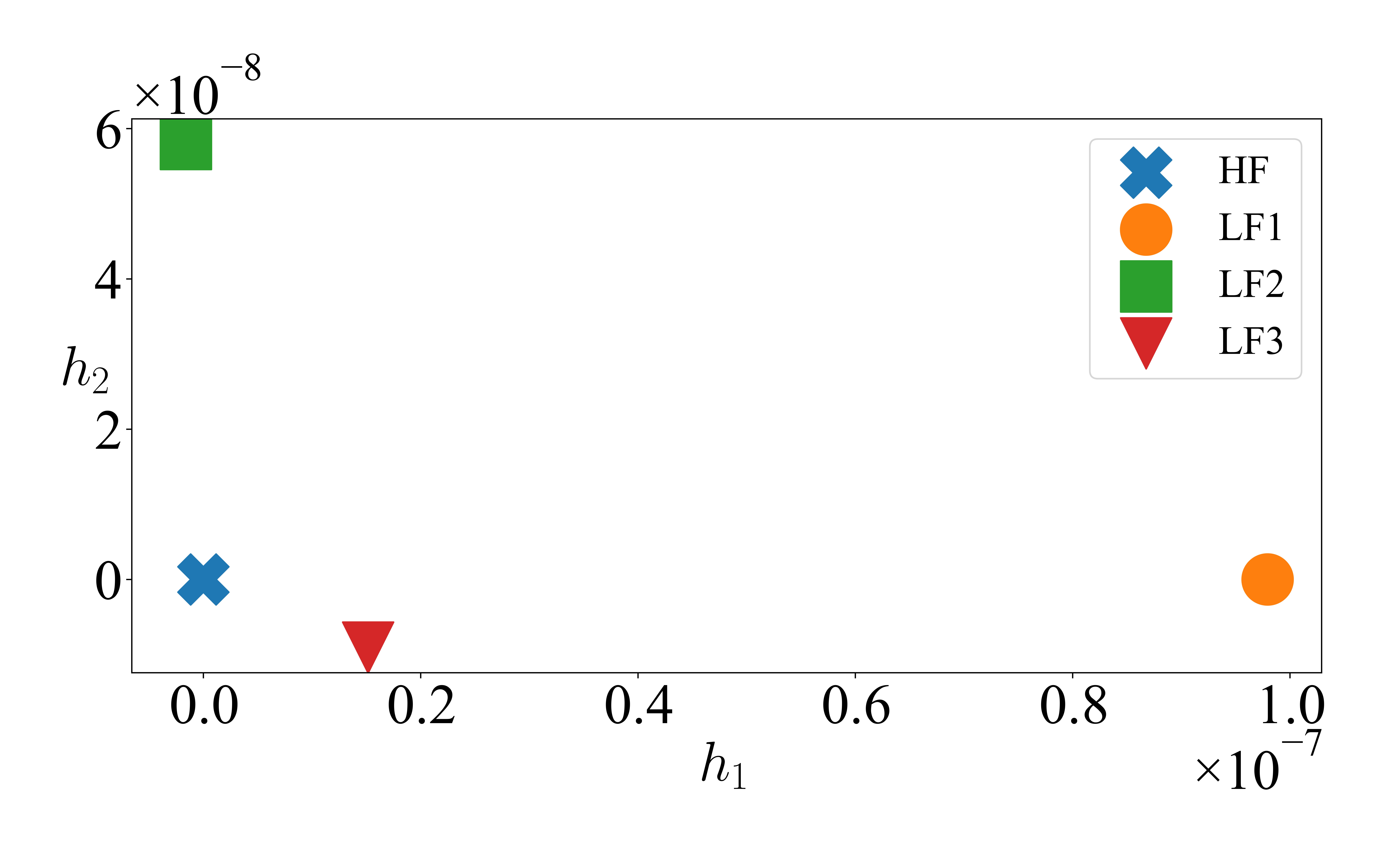}
        \vspace{-9mm}
        \caption{\textbf{\wing}}
    \end{subfigure}%
    \begin{subfigure}{.5\textwidth}
        \centering
        \includegraphics[width=1\linewidth]{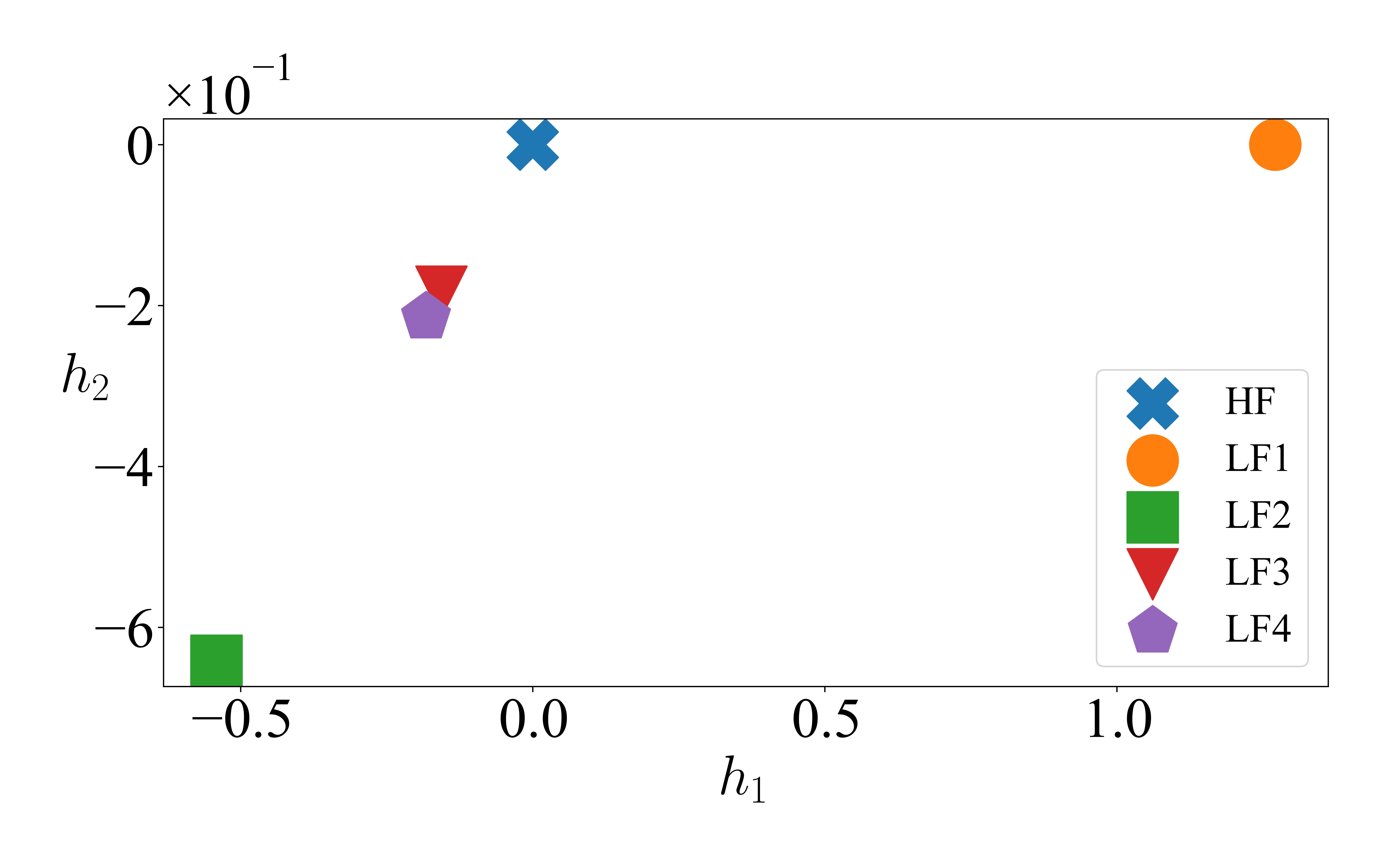}
        \vspace{-9mm}
        \caption{\textbf{\borehole}}
    \end{subfigure}
    \caption{\textbf{Fidelity manifolds of analytic examples:} The plots in \textbf{(a)} and \textbf{(b)} are obtained by fitting an LMGP to the initial data in the \wing~ and \borehole~ examples, respectively. Due to the consistency across the $20$ repetitions, the plots are randomly chosen among them. In \textbf{(b)}, the HF source is encoded far from LF1 and LF2 which indicates that these two sources have large biases with respect to the HF source. \mfbo~excludes these two sources from the BO while \mfbonew~does not.}
    \label{fig: fidelity_plots_analytic}
\end{figure}

\mfbo~excludes the highly biased LF sources from BO before any new samples are obtained (also, during BO, the initial samples from highly biased LF sources are not used in emulation). This exclusion is done based on the latent map of the LMGP model that is trained on the initial data. Figure \ref{fig: fidelity_plots_analytic} shows the latent maps of \wing~ and \borehole~ examples. As shown in Figure \ref{fig: fidelity_plots_analytic}, while all the fidelity sources of \wing~ are beneficial  (since the points encoding the LF sources are very close to the HF point), the first two LF sources of \borehole~ are not correlated enough with the HF (their latent positions are distant from that of the HF) and hence are excluded in \mfbo. However, \mfbonew~ does not require this exclusion because it leverages the biased LF sources merely in the regions that they are correlated with the HF source. In this paper, we do not exclude the biased sources in \mfbo~to better compare it with our proposed method.

\begin{figure}[!b]
    \centering
    \begin{subfigure}{.5\textwidth}
        \centering
        \includegraphics[width=1\linewidth]{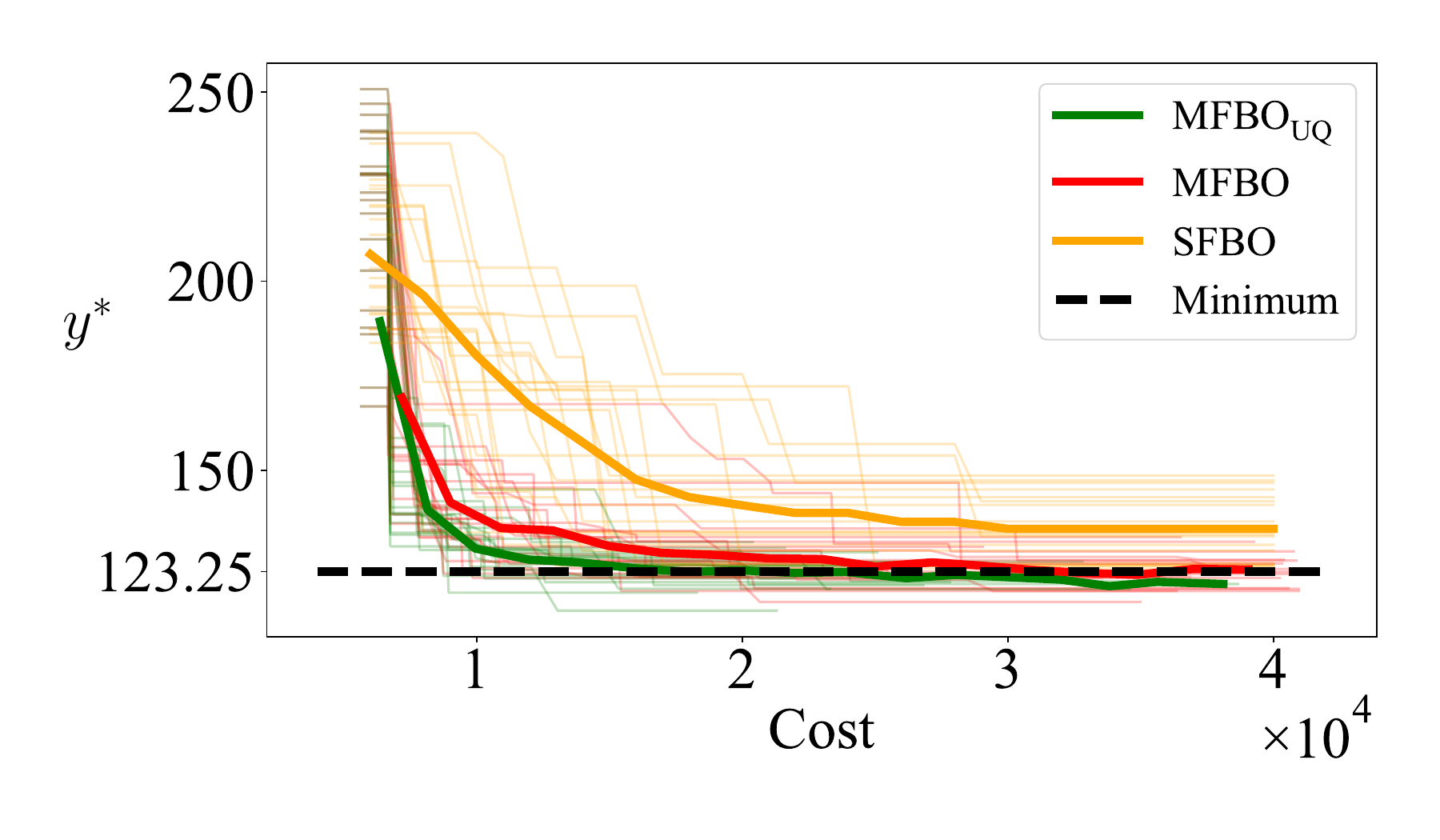}
        \vspace{-9mm}
        \caption{\textbf{\wing}}
        \label{fig: convergence_analytic_wing}
    \end{subfigure}%
    \begin{subfigure}{.5\textwidth}
        \centering
        \includegraphics[width=1\linewidth]{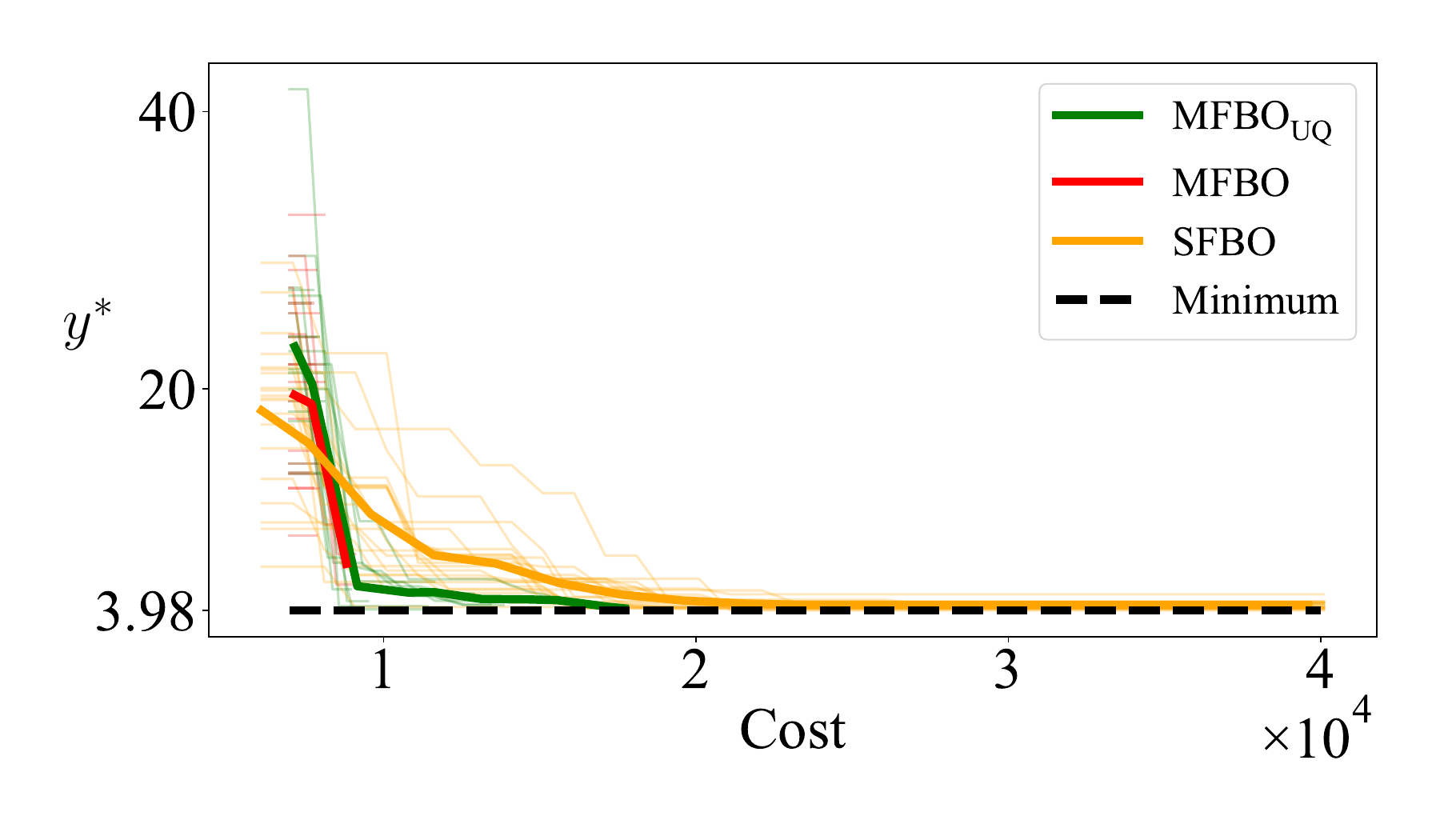}
        \vspace{-9mm}
        \caption{\textbf{\borehole}}
        \label{fig: convergence_analytic_borehole}
    \end{subfigure}
    \caption{\textbf{Convergence histories of analytic examples:} The plots depict the best HF sample found by each approach (\fbest) versus their sampling costs accumulated during the BO iterations (the cost of initial data is included). \textbf{(a)} and \textbf{(b)} summarize the results for the \wing~ and \borehole~ examples, respectively. The thin curves show the convergence history of each repetition and the solid thick ones indicate the average behavior across the $20$ repetitions. In both examples, \mfbonew~ outperforms \mfbo~ in terms of both convergence value and cost. In both examples, SFBO performs the worst. The ground truth is represented by the black dashed line.}
    \label{fig: convergence_analytic}
\end{figure}

Figure \ref{fig: convergence_analytic} summarizes the convergence history of each example by depicting the best HF sample (\fbest) found by each method versus its accumulated sampling cost. As we expect, MF methods (\mfbonew~and \mfbo) outperform SFBO in \wing~(Figure $\ref{fig: convergence_analytic}\textbf{a}$) by leveraging the inexpensive LF sources that are globally correlated with the HF source. However, the superior performance of \mfbonew~ is more obvious in \borehole~ with biased data sources. 

In \borehole~(Figure $\ref{fig: convergence_analytic}\textbf{b}$), all the thin red curves (\mfbo) are straight lines, except for two curves. This means that for $18$ repetitions, the optimization process fails to improve. The reason behind this lack of improvement is that \mfbo~ cannot handle local correlation of the LF sources and samples points that steer the optimization in the wrong direction. Consequently, \mfbo~ cannot find any efficient HF sample with large enough information value (that compensates for its high sampling cost) which results in the lack of improvement. Conversely, all the thin green curves (\mfbonew) converge to a value very close to the ground truth. In addition, \mfbonew~ yields almost the same convergence value as SFBO, but with lower overall computational cost. This instance further demonstrates the effectiveness of our proposed AFs since SFBO is very accurate (albeit more expensive) due to only sampling from HF source and not dealing with the local biases of the LF sources.

\subsection{Real-world Datasets} \label{Sec: real_world_examples}
In this section, we study two materials design problems where the aim is to find the composition that best optimizes the property of interest. We do not add noise to these two examples as they are inherently noisy. The design space of both examples has categorical inputs (denoted by \tb) and we add one more categorical variable (denoted by $s$) to enable data fusion as described in Section \ref{Sec: MF-LMGP}. We design our LMGP to map the categorical inputs onto two $2D$ manifolds (one for \tb~ and the other for $s$) to help with the visualization of the exploration-exploitation behavior of BO in the design space. The HF and LF data are obtained via simulations (based on the density functional theory or DFT) with different fidelity levels.

The first problem is bi-fidelity where the goal is to find the member of the nanolaminate ternary alloy (\nta) family with the largest bulk modulus \cite{yeom2019performance}. The HF and LF datasets have $224$ samples each and are $10$-dimensional ($7$ quantitative and $3$ categorical where the latter have $10$, $12$, and $2$ levels). The cost ratio between the HF and LF sources is $10/1$ and we initialize the BO with $30$ HF and $30$ LF samples (the composition with the largest bulk modulus is never in the initial data). To quantify the robustness of the proposed method to the random initial data, we repeat this process $20$ times for each BO method.

Our second problem is on designing hybrid organic–inorganic perovskite (HOIP) crystals where we aim to find the compound with the smallest inter-molecular binding energy \cite{herbol2018efficient}. In this example, there are $3$ datasets ($1$ from HF and $2$ from LF sources) which have the same dimensionality ($1$ output and $3$ categorical inputs with $10$, $3$, and $16$ levels) but different sizes. The HF dataset has $480$ samples while the first and second LF datasets have $179$ and $240$ samples, respectively. The cost ratio between the three sources is $15/10/5$ (where the HF and LF2 sources are the most expensive and cheapest, respectively) and we initialize the BO with $(15, 20, 15)$ samples for the HF and LF sources (the best compound is excluded from the initial data). We repeat the BO process $20$ times to assess the sensitivity of the results to the initial data. 

\begin{figure}[!b]
    \centering
    \begin{subfigure}{.5\textwidth}
        \centering
        \includegraphics[width=1\linewidth]{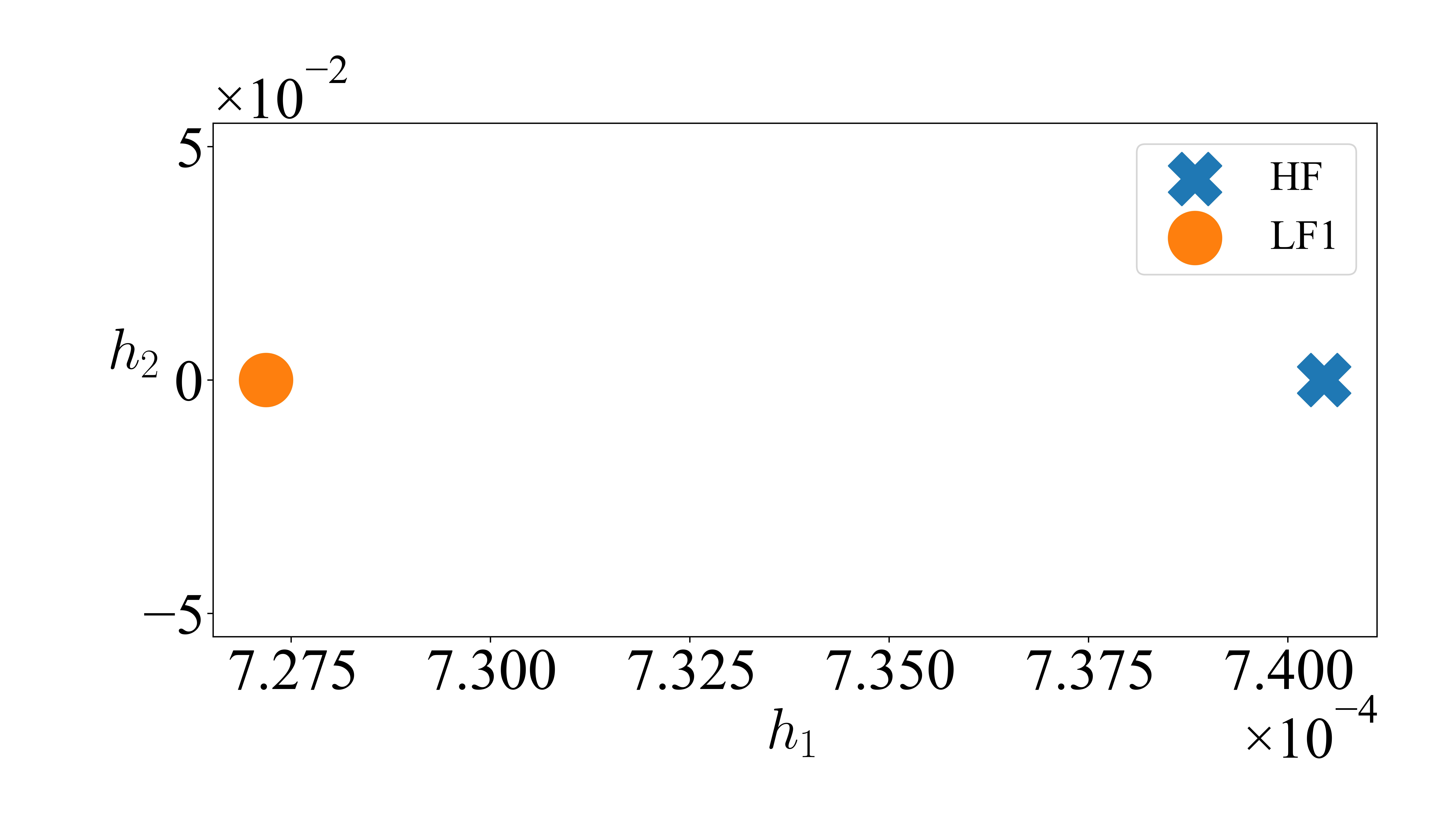}
        \vspace{-9mm}
        \caption{\textbf{\nta}}
        \label{fig: Fidelity_m2ax}
    \end{subfigure}%
    \begin{subfigure}{.5\textwidth}
        \centering
        \includegraphics[width=1\linewidth]{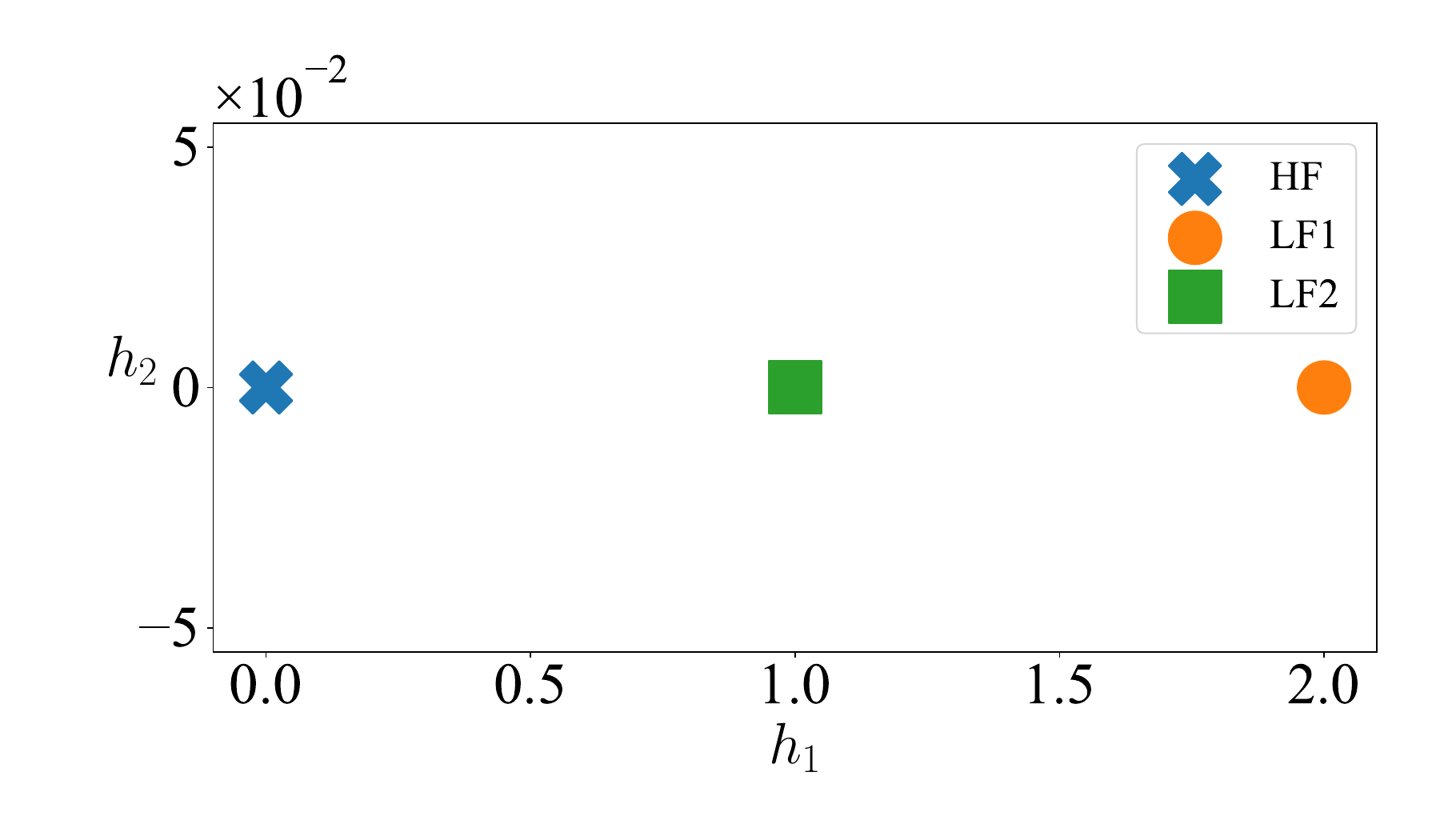}
        \vspace{-9mm}
        \caption{\textbf{\hoip}}
        \label{fig: Fidelity_HOIP}
    \end{subfigure}
    \caption{\textbf{Fidelity manifolds of real-world examples:} The plots in \textbf{(a)} and \textbf{(b)} are obtained by fitting an LMGP to the initial data in the \nta~ and \hoip~ examples, respectively. Due to the consistency across the $20$ repetitions, the plots are randomly chosen among them. In \textbf{(b)}, the HF source is encoded far from LF sources which indicates that these two sources have large biases with respect to the HF source and \mfbo~ should exclude them. However, by excluding these sources, the problem transforms into SF, rendering it incomparable to \mfbonew. To maintain comparability with \mfbonew, we retain the LF sources in \mfbo.}
    \label{fig: fidelity_real_world}
\end{figure}

As mentioned before, the first step in \mfbo~ is to train an LMGP to the initial data in each problem to exclude the highly biased sources. As \nta~ has categorical variables, LMGP learns two manifolds. Based on Figure \ref{fig: fidelity_real_world}\textbf{a}, the latent points of the fidelity sources of \nta~ are very close in the learned fidelity manifold which indicates that there is a high correlation between the corresponding two data sources. However, both latent points of LF sources in \hoip~ are far from the HF one so they both should be excluded due to their large global bias. By excluding both LF sources the MF problem in \hoip reduces to an SF one so we do not exclude the biased LF sources from \hoip~ to be able to compare the performance of \mfbonew~ with \mfbo.

\begin{figure}[!ht]
    \centering
    \begin{subfigure}{.5\textwidth}
        \centering
        \includegraphics[width=1\linewidth]{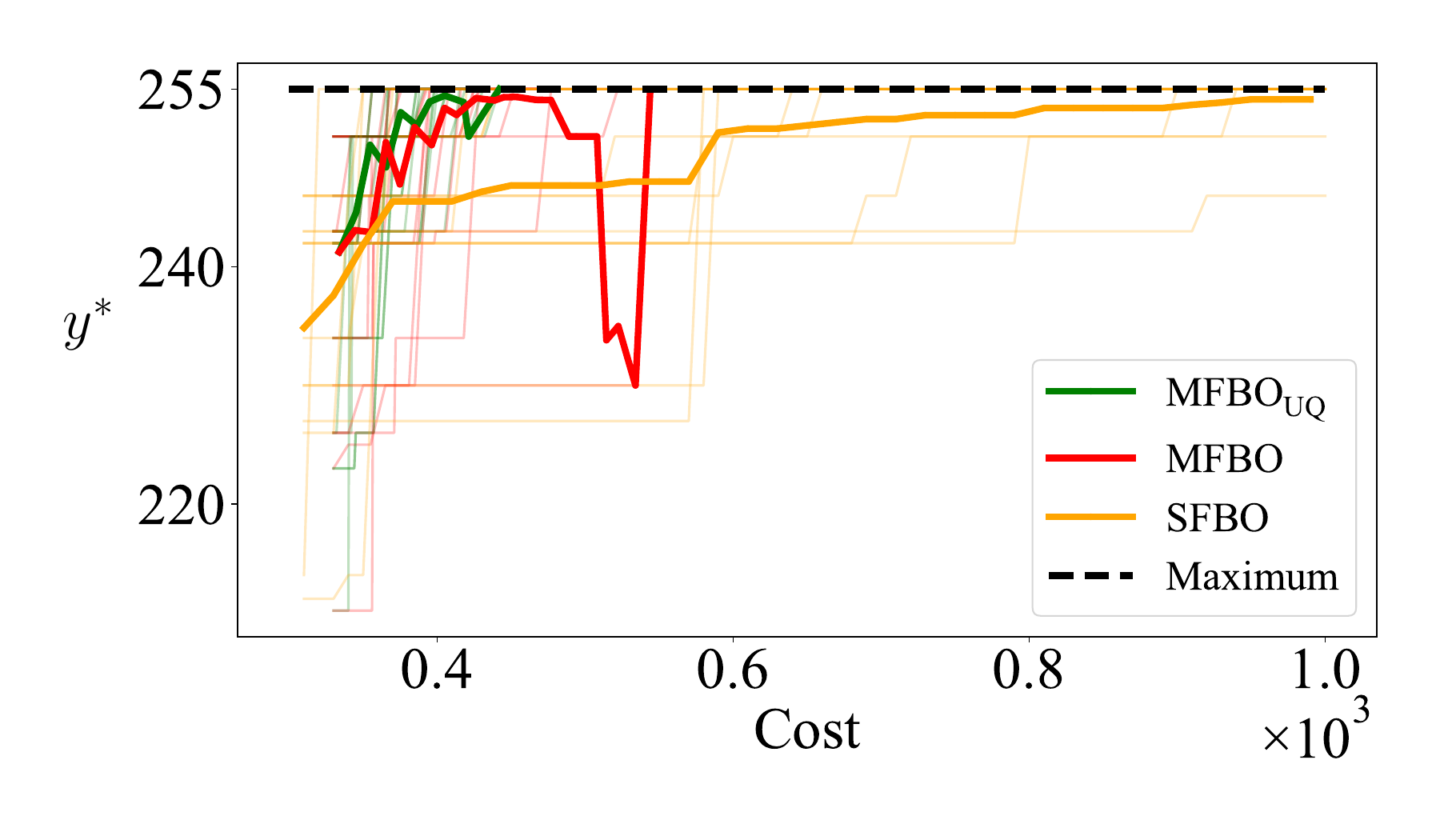}
        \vspace{-9mm}
        \caption{\textbf{\nta}}
        \label{fig: convergence_m2ax}
    \end{subfigure}%
    \begin{subfigure}{.5\textwidth}
        \centering
        \includegraphics[width=1\linewidth]{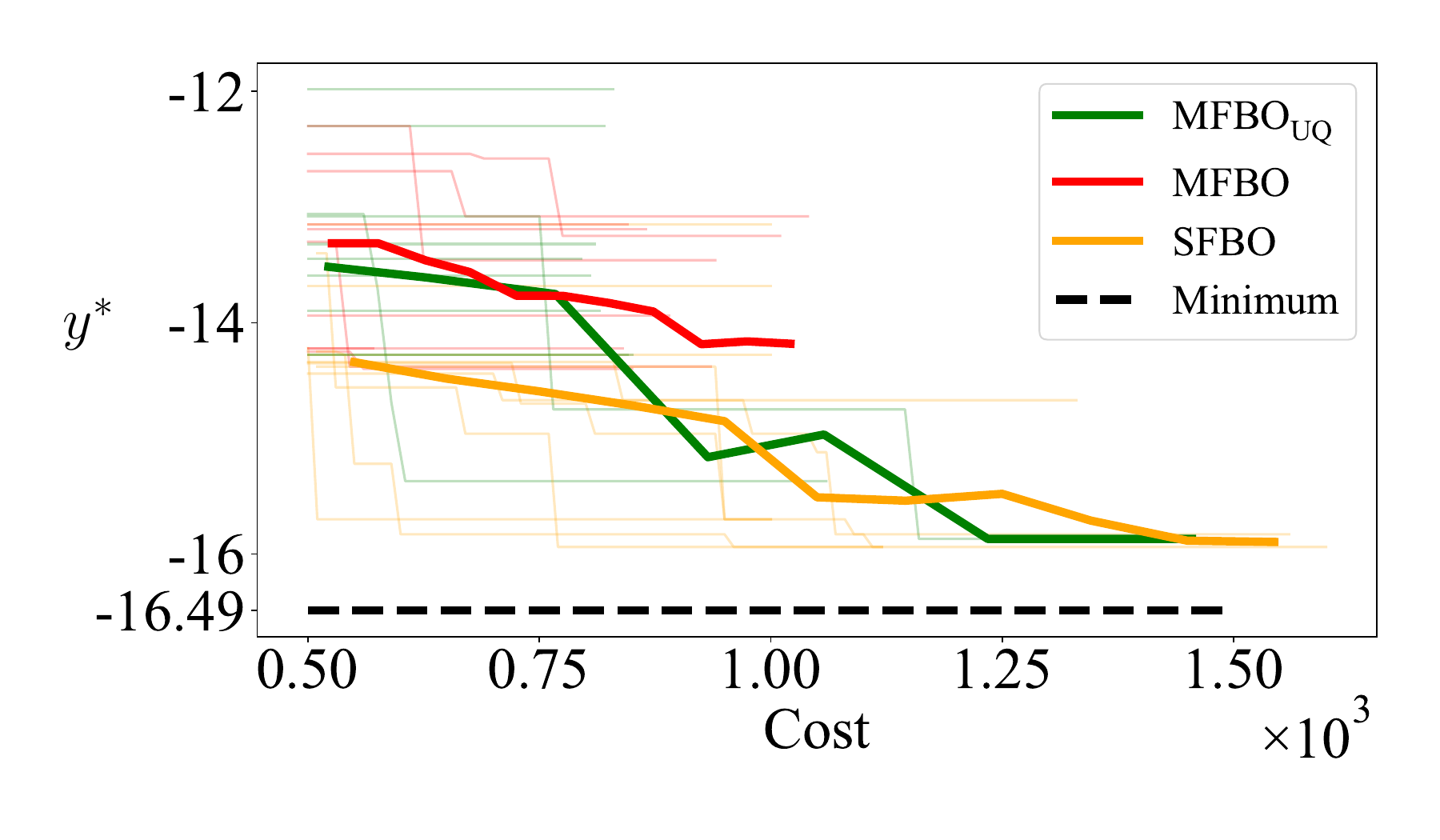}
        \vspace{-9mm}
        \caption{\textbf{\hoip}}
        \label{fig: convergence_HOIP}
    \end{subfigure}
    \caption{\textbf{Convergence histories of real-world examples:} The plots depict the best HF sample found by each approach (\fbest) versus their sampling costs accumulated during the BO iterations (the cost of initial data is included). \textbf{(a)} and \textbf{(b)} summarize the results for the \nta~ and \hoip, respectively. The thin curves show the convergence history of each repetition and the solid thick ones indicate the average behavior across the $20$ repetitions. In \textbf{(b)}, \mfbo~ fails to find the optimum due to it disability in handling biased LF sources. In both examples, \mfbonew~ outperforms other methods.}
    \label{fig: convergence_real_world}
\end{figure}

A summary of the convergence history of \nta~and \hoip~is depicted in Figure \ref{fig: convergence_real_world} by showing the best HF sample (\fbest) found by each method versus its accumulated sampling cost. In Figure \ref{fig: convergence_real_world}\textbf{a}, the LF sources are globally correlated with the HF source and hence both MF methods perform better than SFBO by using inexpensive and informative LF data. Additionally, the higher prediction accuracy of the emulator of \mfbonew~ results in a more efficient sampling and faster convergence of BO in \mfbonew~ compared to \mfbo. 
Regarding the spike in the convergence plot of \mfbo~ in Figure \ref{fig: convergence_real_world}\textbf{a}, we note that $18$ repetitions converge at costs below $500$. Consequently, the thick red line (which is the average across the $20$ repetitions) becomes highly sensitive to the convergence values after cost exceeds $500$ since it is an average of only $2$ values. Specifically, in one of these two repetitions the maximum bulk modulus found is $237$ for many iterations until the cost reaches $544$ when MFBO suddenly converges to the ground truth (i.e., $255$). This sudden convergence results in the spike in the corresponding history and, in turn, the average behavior captured by the thick red line. 

The superiority of \mfbonew~ is more obvious in \hoip~ (see Figure \ref{fig: convergence_real_world}\textbf{b}) which has two highly biased LF sources. In the \hoip~ example, \mfbo~expectedly converges to a sub-optimal compound since both LF sources are only locally correlated with the HF source. So, the AFs fail to sample valuable points to improve the optimization as they cannot find the region where the LF sources are beneficial and informative. Additionally, each data source is obtained from a distinct process so it suffers from different types and levels of noise. Therefore, estimating a single noise for all the data sources in \mfbo~ reduces the emulation accuracy and further exacerbates the performance of AFs. \mfbonew~ overcomes these issues by focusing more on UQ and estimating separate noise processes; resulting in a better performance compared to SFBO and especially to \mfbo.

\begin{figure}[!t]
    \centering
    \begin{subfigure}{.46\textwidth}
        \centering
    \includegraphics[width=1\linewidth]{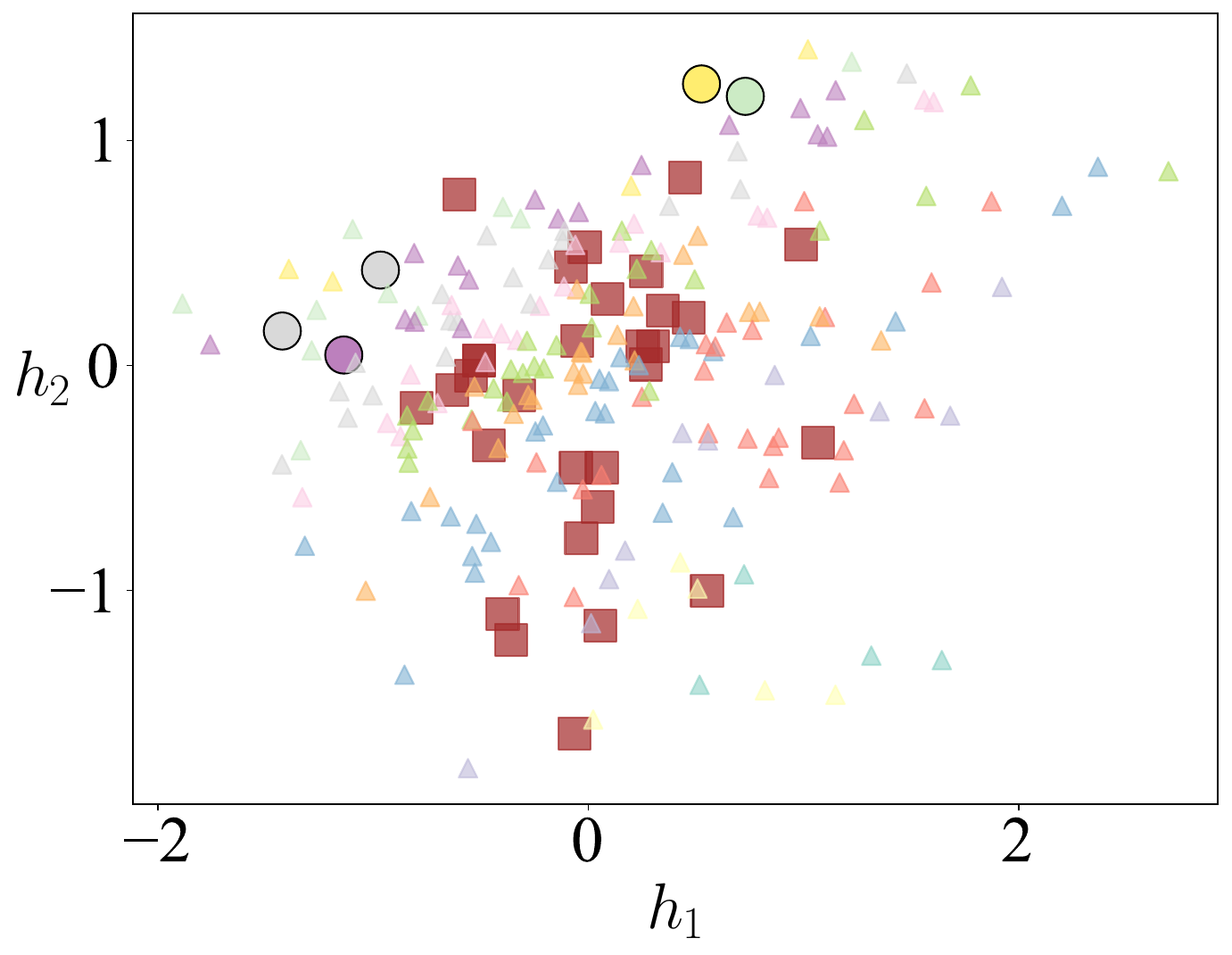}
        \vspace{-7mm}
        \caption{\textbf{HF \mfbo}}
        \label{fig: HF_LM_M2AX_n_multi_n_is}
    \end{subfigure}%
    \begin{subfigure}{.5\textwidth}
        \centering
        \includegraphics[width=1\linewidth]{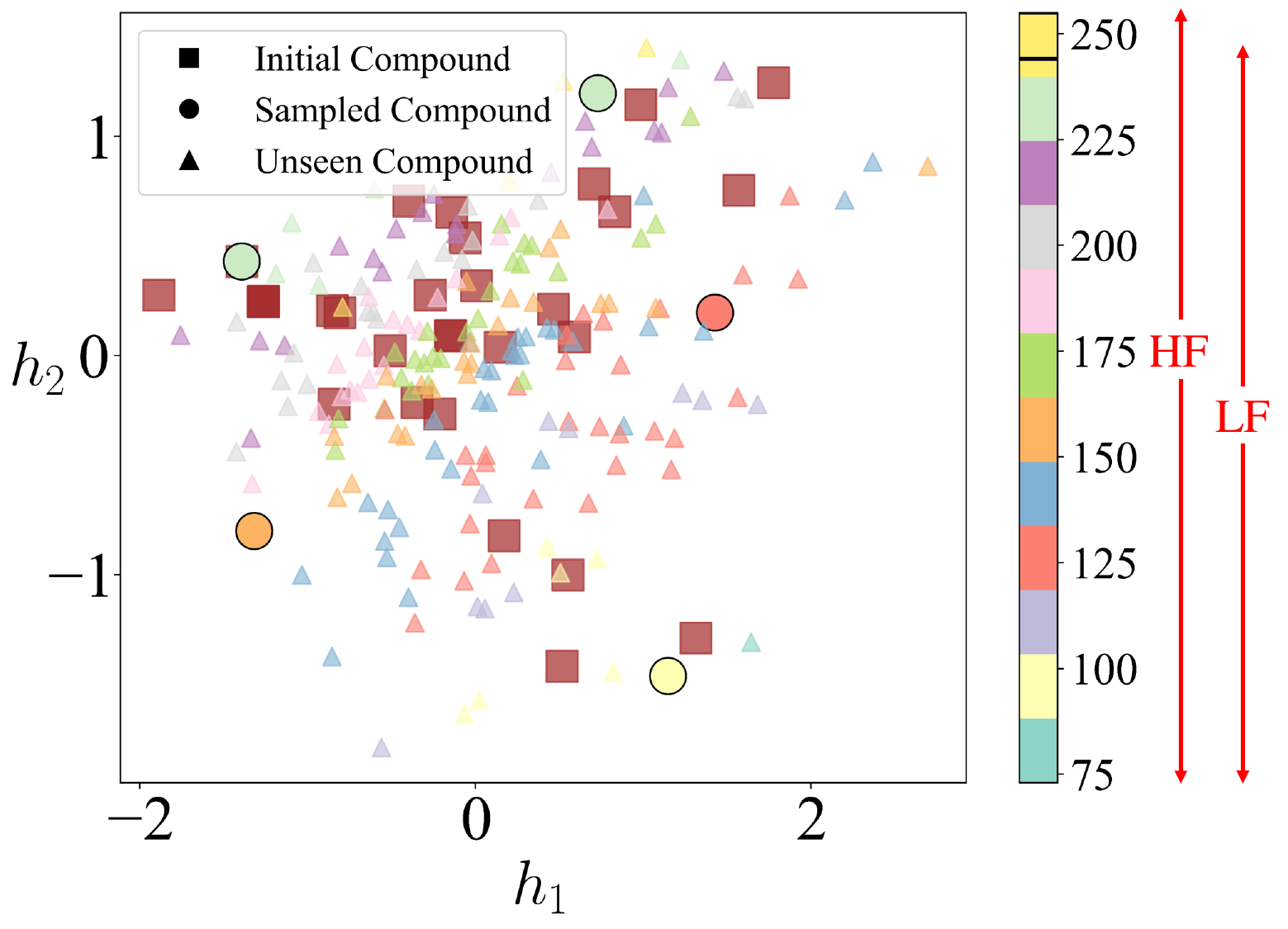}
        \vspace{-7mm}
        \caption{\textbf{LF \mfbo}}
        \label{fig: LF_LM_M2AX_n_multi_n_is}
    \end{subfigure}
    \newline
    \begin{subfigure}{.46\textwidth}
        \centering
        \includegraphics[width=1\linewidth]{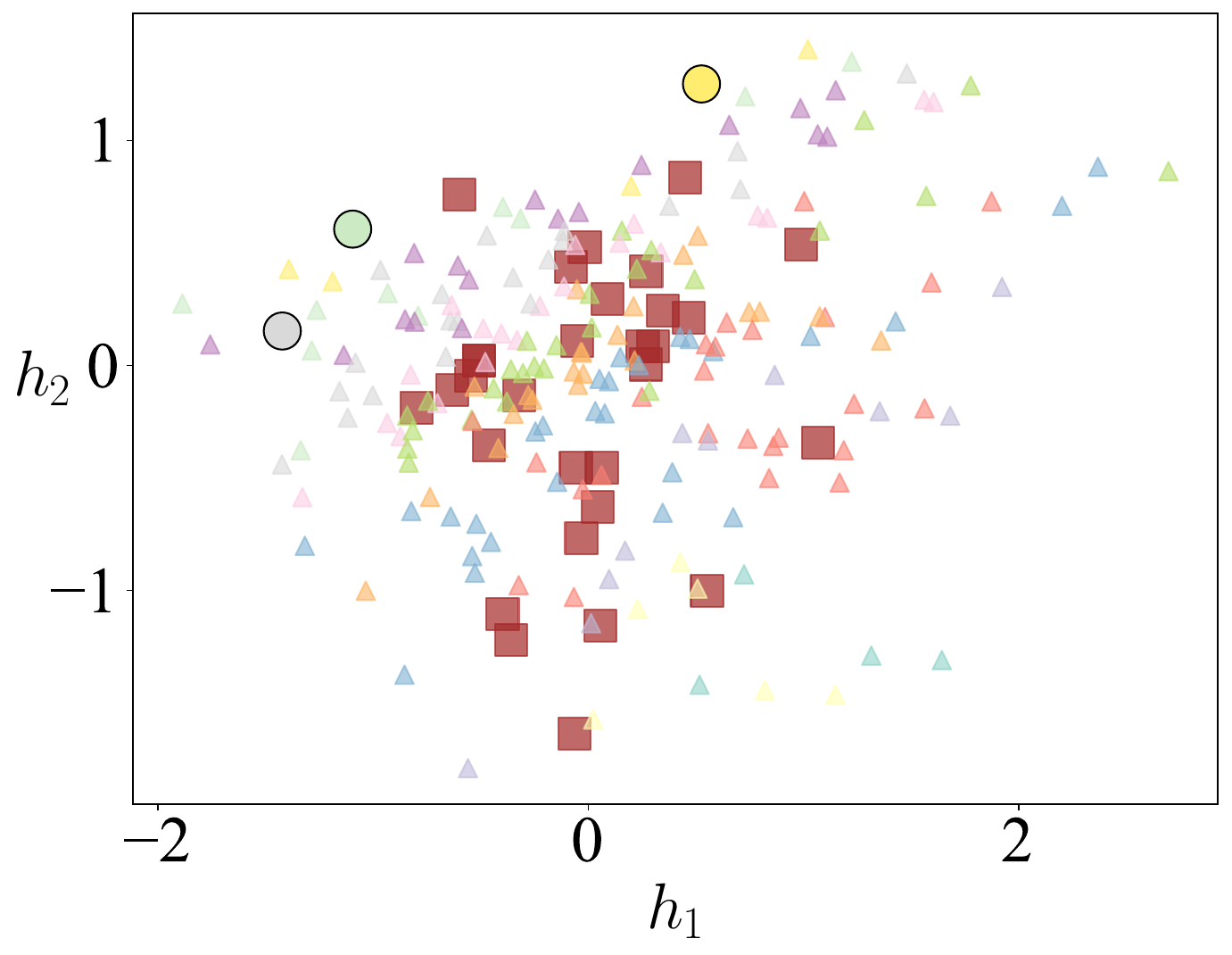}
        \vspace{-7mm}
        \caption{\textbf{HF \mfbonew}}
        \label{fig: HF_LM_M2AX}
    \end{subfigure}
    \begin{subfigure}{.5\textwidth}
        \centering
        \includegraphics[width=1\linewidth]{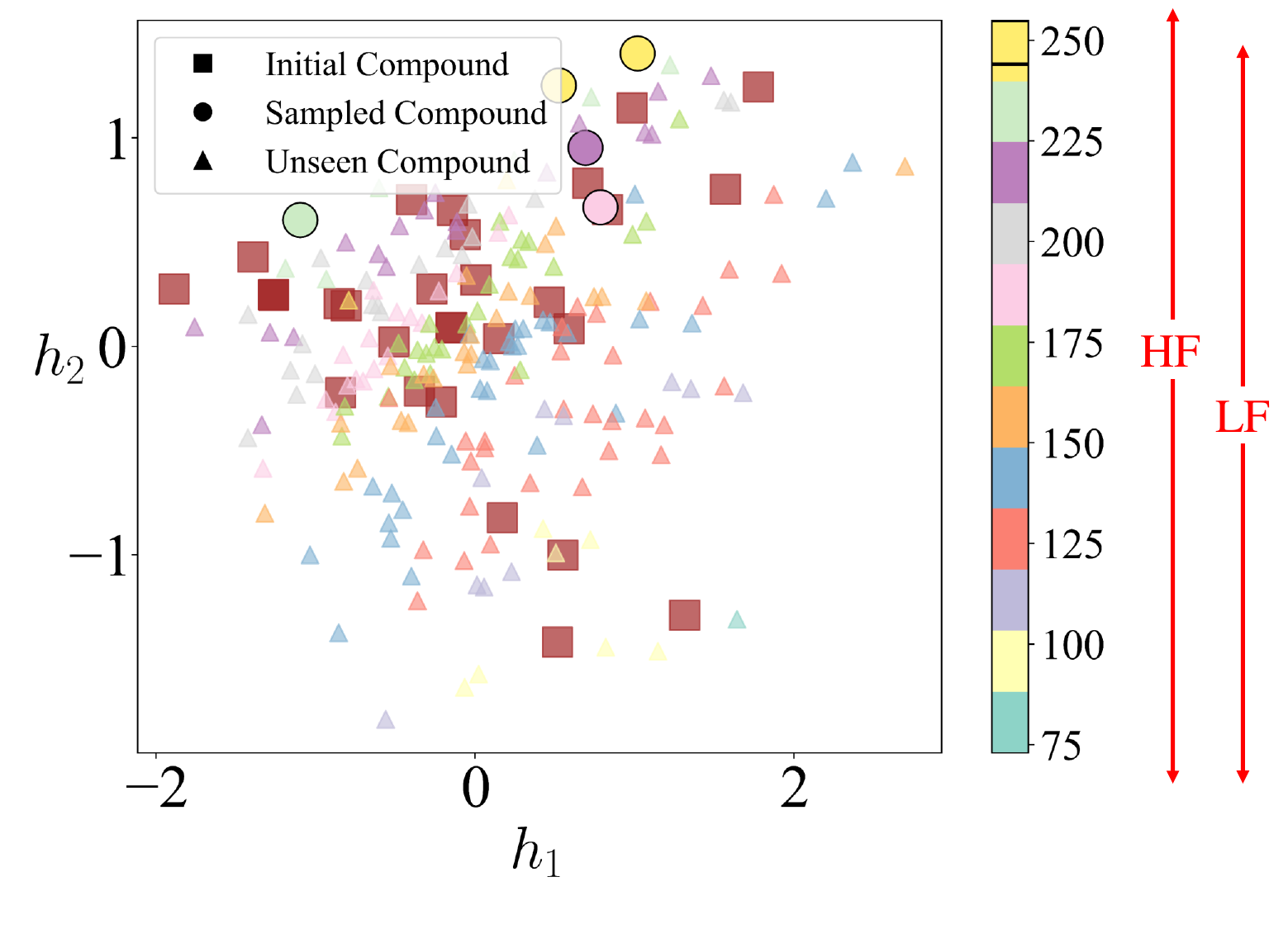}
        \vspace{-7mm}
        \caption{\textbf{LF \mfbonew}}
        \label{fig: LF_LM_M2AX}
    \end{subfigure}
    \caption{\textbf{BO sampling history in the encoded categorical design space of \nta:} The plots in the top and bottom row illustrate the exploration-exploitation behavior of BO in \mfbo~ and \mfbonew, respectively. The left and right columns correspond to the space of HF and LF sources, respectively. All latent points are color-coded based on the ground truth bulk modulus from each source and the marker shapes indicate whether the compound is part of the initial data, sampled during BO, or never seen by LMGP. The red arrows next to the legend indicate the range of response in the two data sources. This figure effectively demonstrates how strategic sampling in \mfbonew~ leads to faster convergence compared to \mfbo (see text for more detailed explanations).}
    \label{fig: trajectory_M2AX}
\end{figure}

\begin{figure}[!b]
  \centering
    \begin{subfigure}{0.3\textwidth}
    \includegraphics[width=\textwidth]{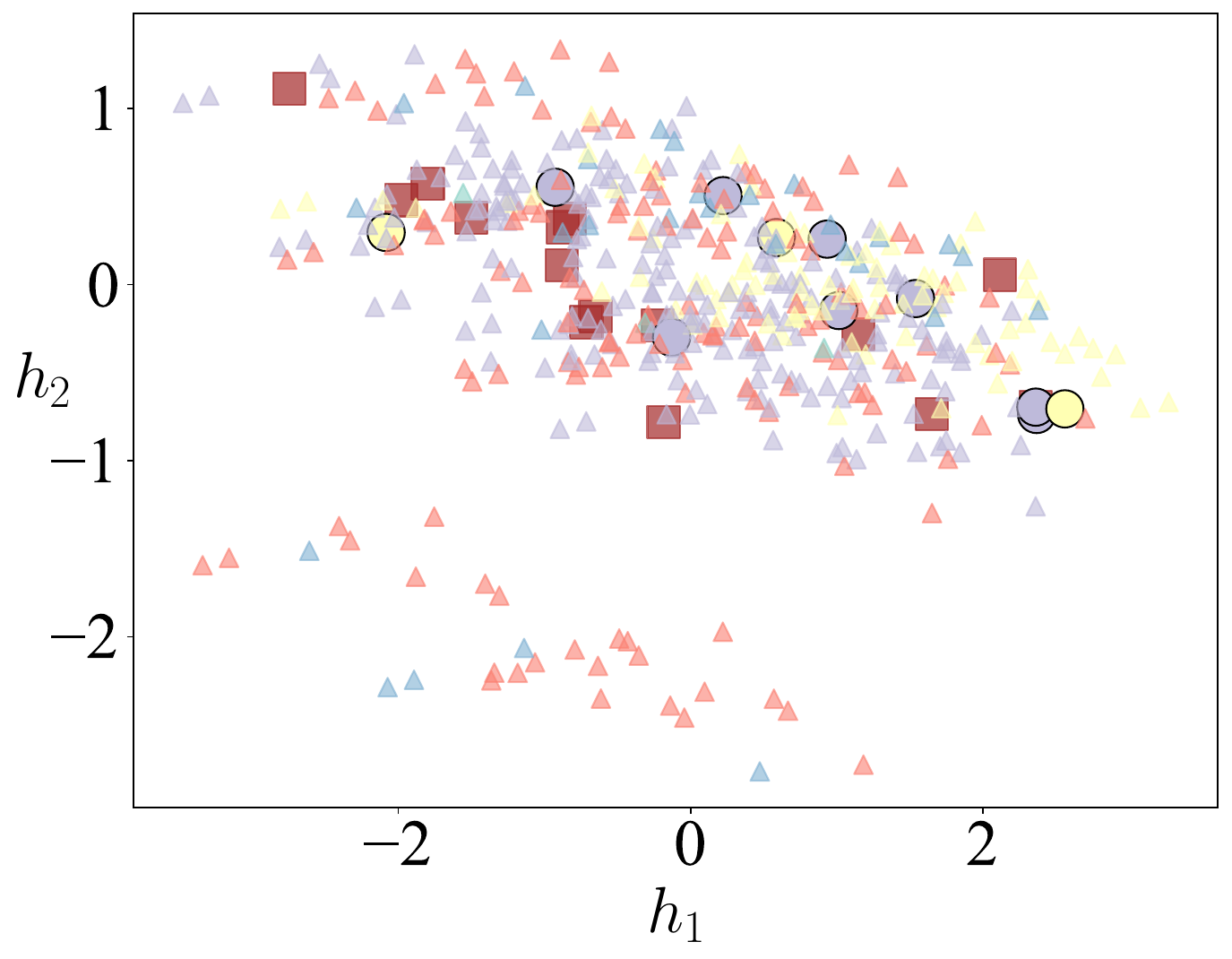}
    \vspace{-7mm}
    \caption{\textbf{HF \mfbo}}
    \label{fig: HF_LM_alpha}
    \end{subfigure}
    \begin{subfigure}{0.3\textwidth}
    \includegraphics[width=\textwidth]{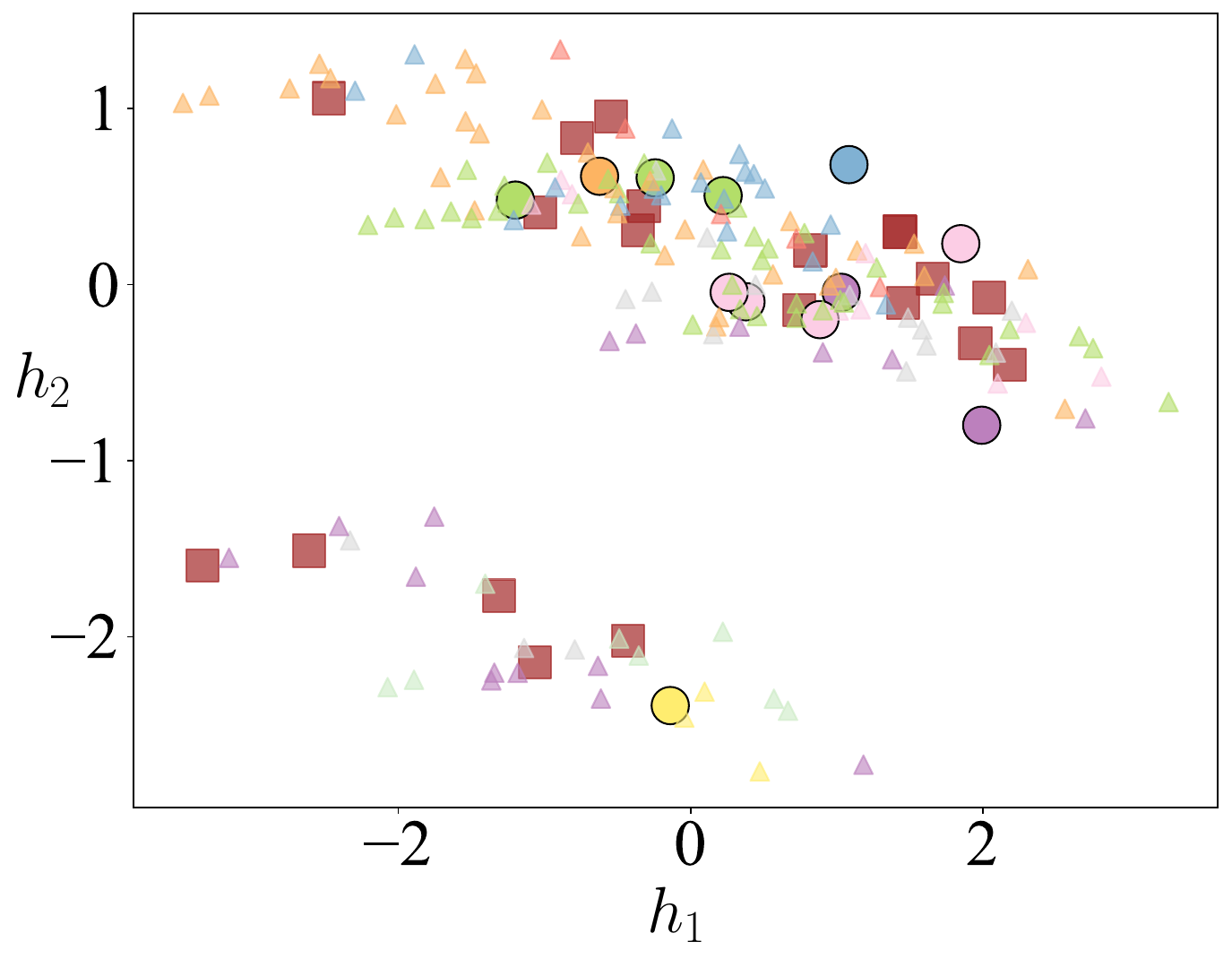}
    \vspace{-7mm}
    \caption{\textbf{LF1 \mfbo}}
    \label{fig: LF1_LM_alpha}
    \end{subfigure}
    \begin{subfigure}{0.36\textwidth}
    \includegraphics[width=\textwidth]{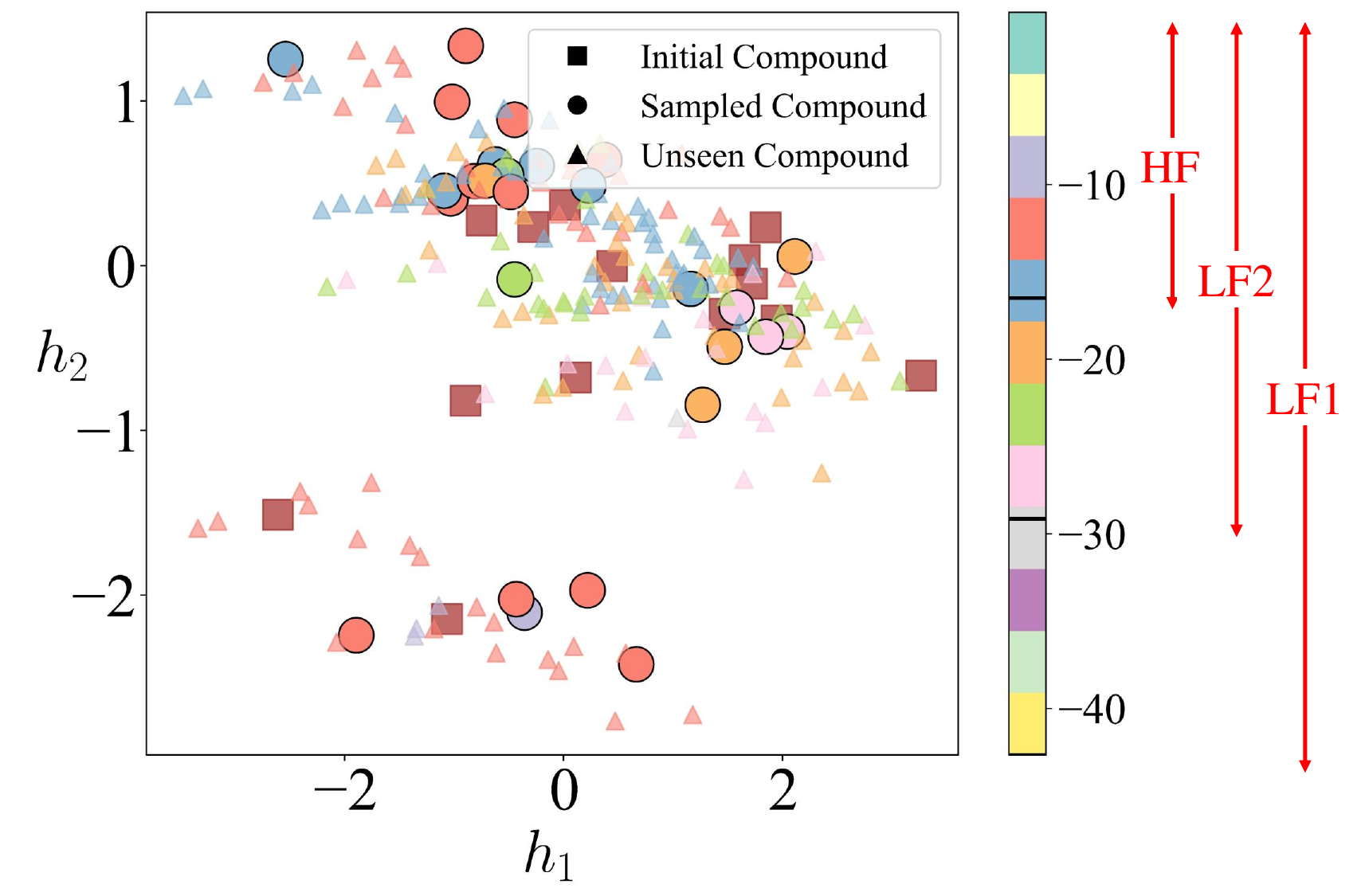}
    \vspace{-7mm}
    \caption{\textbf{LF2 \mfbo}}
    \label{fig: LF2_LM_alpha}
    \end{subfigure}
    \begin{subfigure}{0.3\textwidth}
    \includegraphics[width=\textwidth]{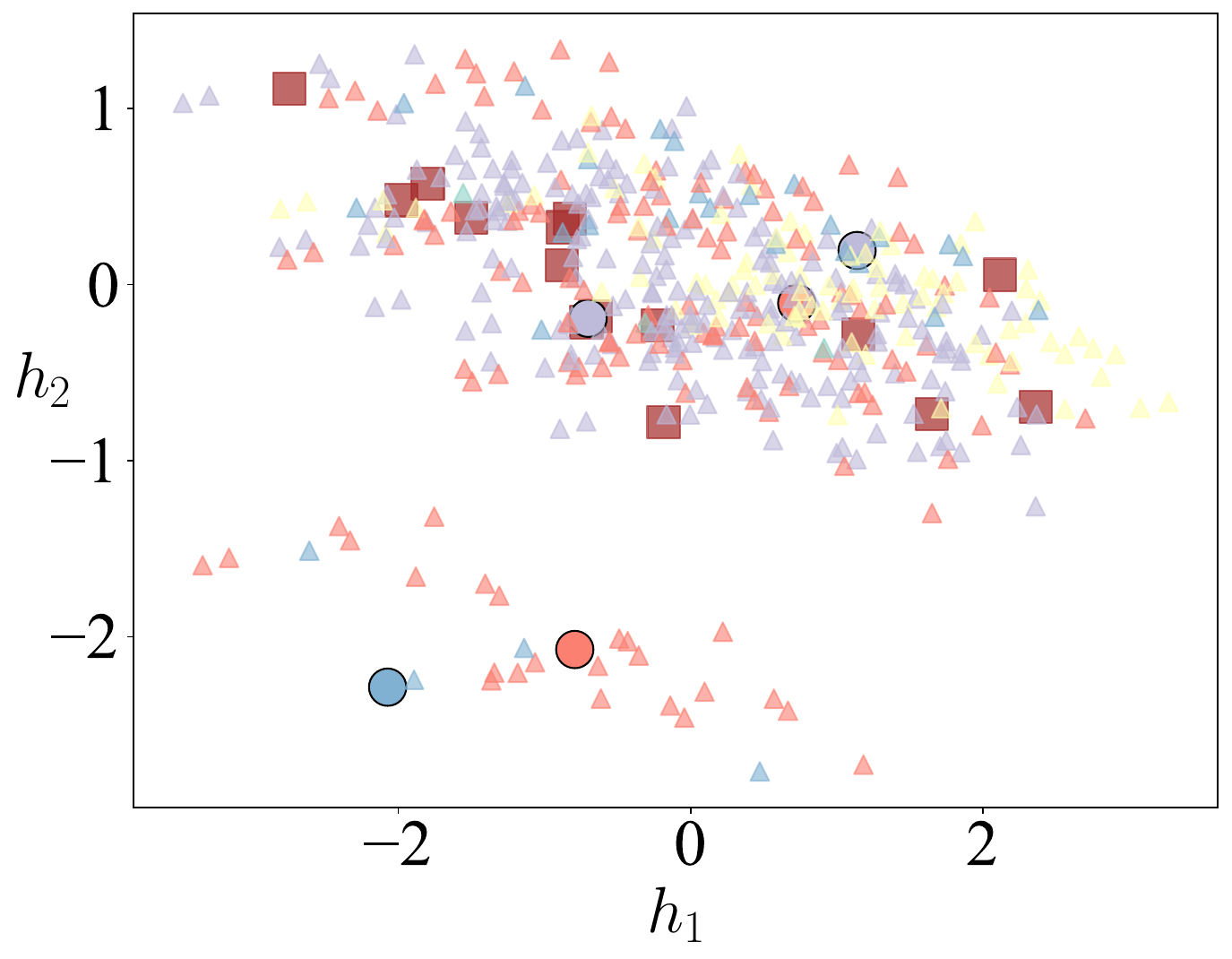}
    \vspace{-7mm}
    \caption{\textbf{HF \mfbonew}}
    \label{fig: HF_LM_beta}
    \end{subfigure}
    \begin{subfigure}{0.3\textwidth}
    \includegraphics[width=\textwidth]{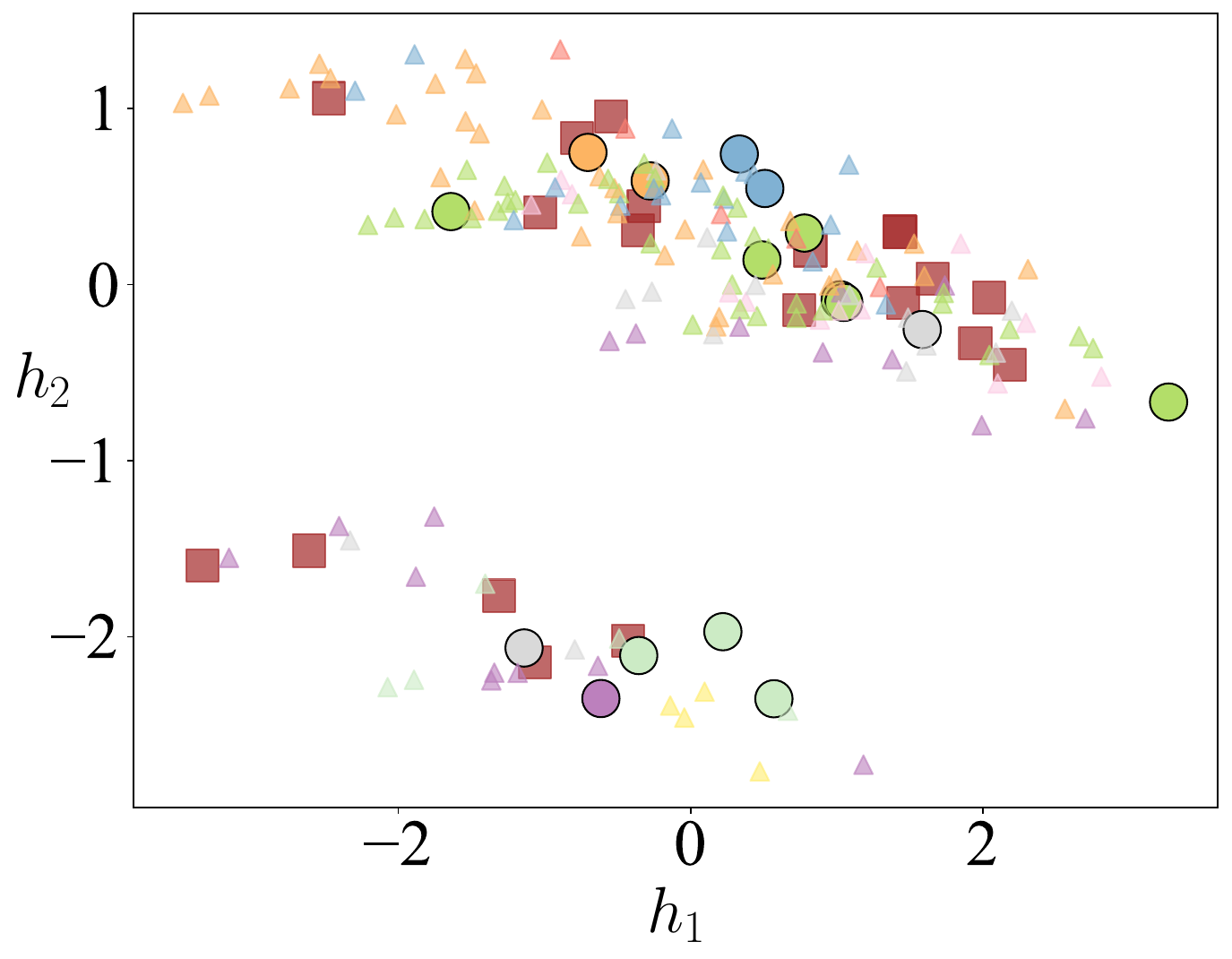}
    \vspace{-7mm}
    \caption{\textbf{LF1 \mfbonew}}
    \label{fig: LF1_LM_beta}
    \end{subfigure}
    \begin{subfigure}{0.38\textwidth}
    \includegraphics[width=\textwidth]{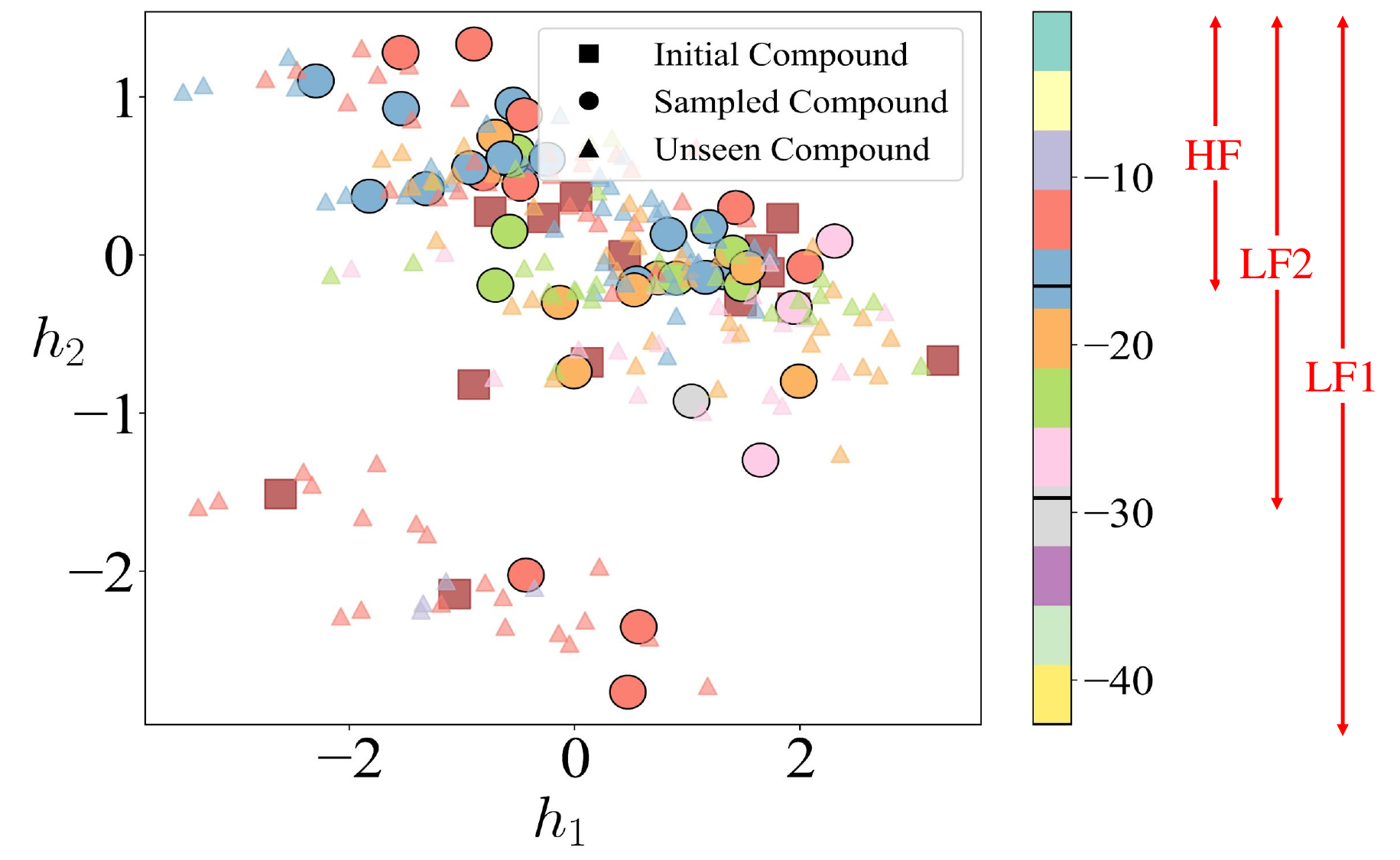}
    \vspace{-7mm}
    \caption{\textbf{LF2 \mfbonew}}
    \label{fig: LF2_LM_beta}
    \end{subfigure}
  \caption{\textbf{BO sampling history in the encoded categorical design space of \hoip:} The plots in the top and bottom row illustrate the exploration-exploitation behavior of BO in \mfbo~ and \mfbonew, respectively. The left, middle and right columns correspond to the space of HF, LF1 and LF2 sources, respectively. All latent points are color-coded based on the ground truth binding energy from each source and the marker shapes indicate whether the compound is part of the initial data, sampled during BO, or never seen by LMGP. The red arrows next to the legend indicate the range of responses in the data sources. This figure demonstrates how the strategic sampling in \mfbonew~enables it to find the optimum while \mfbo~ fails. }
  \label{fig: trajectory_HOIP}
\end{figure}

The $2D$ manifolds in Figures \ref{fig: trajectory_M2AX} and \ref{fig: trajectory_HOIP} demonstrate the trajectory of BO in the categorical design space of each data source in \nta~ and \hoip, respectively. The top and bottom rows of these figures correspond to \mfbo~ and \mfbonew. In these manifolds, each latent point indicates a compound and is color coded based on the ground truth response value (i.e., the bulk modulus) from each source. The marker shapes in these manifolds indicate whether a compound is part of the initial data, sampled during BO, or never seen by LMGP. As expected, most markers are triangles which indicates that most combinations are never tested by either \mfbo~ or \mfbonew. The red arrows next to the legend mark the range of response in the data sources which indicate that, unlike in Figure \ref{fig: trajectory_M2AX} for \nta, the response ranges across the three sources are quite different in the \hoip~problem. 

To benefit any MFBO approach, LF sources should be sampled in two primary regions of their input space: $(1)$ the region that contains their own optima since each data source is analyzed separately in the auxiliary optimization problems (see Section \ref{Sec: CA-AF} for details), and $(2)$ the region where the LF sources are correlated with the HF source. These two regions may overlap with each other (as is the case in \nta) or not (as is the case in \hoip~ or the $1D$ example in Figure \ref{fig: 1d}\textbf{b} where \mfbonew~only samples LF2 once when $x<5$). 
We note that exploring the correlation region (if it exists!) is crucial for capturing the relationship between the LF and HF sources and as shown below the effectiveness of this exploration highly depends on the accuracy of the emulator in surrogating each source, estimating uncertainties, and identifying the correlation patterns among different data sources. 


As shown in Figure \ref{fig: trajectory_M2AX}, for both \mfbo~and \mfbonew~manifolds with very similar structures are learnt by LMGP for HF and LF data (this was expected per Figure \ref{fig: fidelity_real_world} which indicates that the two sources are highly correlated). For instance, for both LF and HF data, the optimum compound is located at the top-right corner of the manifold and their values are also quite close ($255$ for HF and $244$ for LF). This similarity indicates that \mfbo~and \mfbonew~are both able to learn about the HF source by sampling the space of the LF source. However, this sampling is more effective in the case of \mfbonew~ since its emulator quantifies the uncertainties more accurately. In particular, \mfbonew~correctly samples compounds from the LF source that are mostly encoded in the top right corner of the manifold (see Figure \ref{fig: trajectory_M2AX}\textbf{d}) while \mfbo~tests compounds that explore the entire design space (see Figure \ref{fig: trajectory_M2AX}\textbf{b}). 



As shown in Figure \ref{fig: trajectory_HOIP}, for any of the sources and with either \mfbo~ or \mfbonew, the compounds in the \hoip~example are encoded by LMGP into two major clusters where the smaller one contains the optimum design. By examining these two clusters we observe that all the compounds in the smaller cluster have Dimethylformamide (DMF) solvent. These observations are quite interesting in that they provide engineers with insights into the most important design variables that affect the materials properties (e.g., DMF solvent which decreases the binding energy in this example). 

The initial HF dataset used in either \mfbo~ or \mfbonew~ (see Figures \ref{fig: trajectory_HOIP}\textbf{(a)} and \ref{fig: trajectory_HOIP}\textbf{(d)}) is very small and does not have any compounds from the small cluster that contains the optimum. However, there are some initial samples from LF1 and LF2 in this cluster and so we should expect BO to leverage these samples (and the fact that they have some correlation with the unseen HF compounds) in emulating the HF source and sampling compounds from it that belong to the small cluster. While this expectation is met by \mfbonew, \mfbo~ fails to explore the (encoded) design space that contains the optimum HF sample. This failure is because $(1)$ both LF sources (especially LF1) provide smaller binding energies than the HF source, and $(2)$ the emulator of \mfbo~overestimates the uncertainties in LF sources. The combination of these two factors prevents \mfbo~to find an HF sample that is valuable enough to be selected in Eq. \ref{eq: auxiliary-opt}.

    \section{Conclusion} \label{Sec: conclusion}
In this paper, we develop a novel method to improve the performance of multi-fidelity cost-aware BO techniques. Our method enhances the accuracy and convergence rate of MFBO through two main contributions. Firstly, we enable the emulator to estimate separate noise processes for each source of data. This feature increases the accuracy of the trained model since different data sources may exhibit different types and levels of noise. Secondly, we define a new objective function penalized by strictly proper scoring rules to $(1)$ improve the prediction, $(2)$ increase the focus on UQ, and $(3)$ forgo the need to exclude highly biased data sources from BO. 
Our BO method, \mfbonew, accommodates any number of data sources with any levels of noise, does not require any prior knowledge about the relative accuracy of (or relation between) these sources, and can handle both continuous and categorical variables. In this paper, we illustrate these features via both analytic and engineering problems.


In this work, we use two fixed AFs in each iteration. However, one can also customize the choice of AFs for different iterations using adaptive approaches. Additionally, the examples presented in this paper are limited to single-objective problems and we do not aim to exclude the effect of noise in the final solution (i.e., the best HF sample found is noisy). We intent to study these direction in our future works.

\section*{Acknowledgments}
We appreciate the support from National Science Foundation (award number $CMMI-2238038$), the Early Career Faculty grant from NASA’s Space Technology Research Grants Program (award number $80NSSC21K1809$), and the UC National Laboratory Fees Research Program of the University of California (Grant Number $L22CR4520$). 
    \appendix
\addcontentsline{toc}{section}{Appendices}
\section*{Appendices}

\section{Table of Numerical Examples} \label{sec: appendix-table}
\setcounter{equation}{0}
\renewcommand{\theequation}{\thesection-\arabic{equation}}

\begin{table}[!ht]
    \setlength{\extrarowheight}{0pt}
    \setlength{\tabcolsep}{5pt} 
    {\renewcommand{\arraystretch}{2}
    \centering
    \begin{tabular}{
        |c|
        m{0.1\linewidth}
        |c|
        m{0.1\linewidth}
        |c|
        m{0.06\linewidth}|
    }
        \hline
        \textbf{Name} &
        \textbf{Source ID} &
        \textbf{Formulation} &
        \textbf{n}  &
        \textbf{RRMSE} &
        \textbf{Cost} \\
        \hline
        \multirow{5}{*}{\textbf{Borehole}} &
        HF &
        $\frac{2 \pi T_{u}(H_u-H_l)}{\ln (\frac{r}{r_w})(1+\frac{2 L T_u}{\ln (\frac{r}{r w}) r_w^2 k_w}+\frac{T_{u}}{T_l})}$ &
        $5$ &
        $-$ &
        $1000$ \\
        \hhline{~|-----|}
        & LF1 &
        $\frac{2 \pi  T_{\mathrm{u}} (H_{\mathrm{u}}-0.8 H_l)} {\ln (\frac{r}{r_w})(1+\frac{1 L T_{\mathrm{u}}}{\ln (\frac{r}{r_w}) r_w^2 k_w}+\frac{T_{u}}{T_l})}$ &
        $5$ &
        $4.40$ &
        $100$ \\
        \hhline{~|-----|}
        & LF2 &
        $\frac{2 \pi T_u(H_u- H_l)}{\ln (\frac{r}{r_w})(1+\frac{8 L T_u}{\ln (\frac{r}{r w}) r_w^2 k_w}+0.75 \frac{T_{u}}{T_l})}$ &
        $50$ &
        $1.54$ &
        $10$ \\
        \hhline{~|-----|}
        & LF3 &
        $\frac{2 \pi T_{\mathrm{u}}(1.09 H_u-H_l)}{\ln (\frac{4 r}{r_w})(1+\frac{3 L T_{u}}{\ln (\frac{r}{r_w}) r_w^2 k_w}+\frac{T_u}{T_l})}$ &
        $5$ &
        $1.30$ &
        $100$ \\
        \hhline{~|-----|}
        & LF4 &
        \makecell{%
          $\frac{2 \pi T_{\mathrm{u}}(1.05 H_u-H_l)}{\ln (\frac{2 r}{r_w})}$ \\ [0.5ex]
          $\times \left(1+\frac{3 L T_{i u}}{\ln (\frac{r}{\tau w}) r_w^2 k_{\mathrm{W}}}+ \frac{T_{\mathrm{u}}}{T_l}\right)$
        } &
        $50$ &
        $1.3$ &
        $10$ \\
        \hline
        \multirow{4}{*}{\textbf{Wing}} &
        HF &
        \makecell{%
          $0.036 s_w^{0.758} w_{f w}^{0.0035}(\frac{A}{\cos ^2(\Lambda)})^{0.6}$ \\ [0.5ex]
          $\times \lambda^{0.04}(\frac{100 t_c}{\cos (\Lambda)})^{-0.3}(N_z W_{d g})^{0.49}+s_w w_p$
        } &
        $5$ &
        $-$ &
        $1000$ \\
        \hhline{~|-----|}
        & LF1 &
        \makecell{%
          $0.036 s_w^{0.758} w_{f w}^{0.0035}(\frac{A}{\cos ^2(\Lambda)})^{0.6}$ \\ [0.5ex]
          $\times \lambda^{0.04}(\frac{100 t_c}{\cos (\Lambda)})^{-0.3}(N_z W_{d g})^{0.49}+w_p$
        } &
        $5$ &
        $0.19$ &
        $100$ \\
        \hhline{~|-----|}
        & LF2 &
        \makecell{%
          $0.036 s_w^{0.8} w_{f w}^{0.0035}(\frac{A}{\cos ^2(\Lambda)})^{0.6}$ \\ [0.5ex]
          $\times \lambda^{0.04}(\frac{100 t_c}{\cos (\Lambda)})^{-0.3}(N_z W_{d g})^{0.49}+w_p$
        } &
        $10$ &
        $1.14$ &
        $10$ \\
        \hhline{~|-----|}
        & LF3 &
        \makecell{%
          $0.036 s_w^{0.9} w_{f w}^{0.0035}(\frac{A}{\cos ^2(\Lambda)})^{0.6}$ \\ [0.5ex]
          $\times \lambda^{0.04}(\frac{100 t_c}{\cos (\Lambda)})^{-0.3}(N_z W_{d g})^{0.49}$
        } &
        $50$ &
        $5.75$ &
        $1$ \\
        \hline
    \end{tabular} 
    }
    \caption{\textbf{List of analytic functions}: $n$ denotes the number of initial samples. The relative root mean squared error (RRMSE) of an LF source is calculated by comparing its output to that of the HF source at $10000$ random points. The cost column is the cost of obtaining a sample from the corresponding source.}
    \label{table: analytic-formulation}
\end{table}

    \pagebreak    
    \printbibliography
\end{document}